\documentclass[lettersize,journal]{IEEEtran}
\usepackage[colorlinks=true, linkcolor=blue, citecolor=blue, urlcolor=blue]{hyperref}
\usepackage{cite}
\usepackage{amsmath,amssymb,amsfonts}
\usepackage{algorithmic}
\usepackage{graphicx}
\usepackage{textcomp}
\usepackage{xcolor}
\usepackage{colortbl}
\usepackage{multirow}
\usepackage{multicol}
\usepackage{arydshln}
\usepackage{booktabs}
\usepackage{xspace}
\usepackage{subfigure}

\newcommand{\eg}{\textit{e.g.}\xspace}

\newcommand{\bluedown}[1]{$_{\color{blue}\downarrow #1}$}
\newcommand{\redup}[1]{$_{\color{red}\uparrow #1}$}

\begin{document}

\title{MIFNet: Learning Modality-Invariant Features for Generalizable Multimodal  Image Matching}

\author{Yepeng Liu, Zhichao Sun, Baosheng Yu, Yitian Zhao, \IEEEmembership{Senior Member, IEEE},\\  Bo Du, \IEEEmembership{Senior Member, IEEE},  Yongchao Xu, \IEEEmembership{Member, IEEE}, and Jun Cheng, \IEEEmembership{Senior Member, IEEE}
\thanks{Y. Liu, Z. Sun, B. Du and Y. Xu are with National Engineering Research Center for Multimedia Software, Institute of Artificial Intelligence, School of Computer Science, and Hubei Key Laboratory of Multimedia and Network Communication Engineering, Wuhan University, Wuhan, 430070, China. (E-mail: \{yepeng.liu, zhichaosun, dubo, yongchao.xu\}@whu.edu.cn). (Corresponding author: Yongchao Xu.)} 
\thanks{Baosheng Yu is with the Lee Kong Chian School of Medicine, Nanyang Technological University, 308232, Singapore. (e-mail: baosheng.yu@ntu.edu.sg).}
\thanks{Yitian Zhao is with the Ningbo Institute of Materials Technology and Engineering, Chinese Academy of Sciences, Ningbo, Zhejiang 315211, China
(e-mail: yitian.zhao@nimte.ac.cn).}
\thanks{ Jun Cheng is with the Institute for Infocomm Research (I$^2$R),
Agency for Science, Technology and Research (A*STAR), 1 Fusionpolis Way, \#21-01, Connexis South Tower, Singapore 138632, Republic of Singapore
(e-mail: cheng\_jun@i2r.a-star.edu.sg).}}
\maketitle

\markboth{$>$ \normalsize{A Submission to IEEE Transactions on Image Processing} $<$}%
{Shell \MakeLowercase{\textit{et al.}}: A Sample Article Using IEEEtran.cls for IEEE Journals}


\maketitle
\begin{abstract}
Many keypoint detection and description methods have been proposed for  image matching or registration. 
While these methods demonstrate promising performance for single-modality image matching, they often struggle with multimodal data because the descriptors trained on single-modality data tend to lack robustness against the non-linear variations present in multimodal data. 
Extending such methods to multimodal image matching often requires well-aligned multimodal data to learn modality-invariant descriptors. However, acquiring such data is often costly and impractical in many real-world scenarios. 
To address this challenge, we propose a modality-invariant feature learning network (MIFNet) to compute modality-invariant features for keypoint descriptions in multimodal image matching using only single-modality training data. Specifically, we propose a novel latent feature aggregation module and a cumulative hybrid aggregation module  to enhance the base keypoint descriptors trained on single-modality data by leveraging pre-trained features from Stable Diffusion models. 
We validate our method with recent keypoint detection and description methods in three multimodal retinal image datasets (CF-FA, CF-OCT, EMA-OCTA) and two remote sensing datasets (Optical-SAR and Optical-NIR).  
Extensive experiments demonstrate that the proposed MIFNet is able  to learn modality-invariant feature  for multimodal image matching without accessing the targeted modality and  has good zero-shot generalization ability. The code will be released at  https://github.com/lyp-deeplearning/MIFNet.
\end{abstract}





\section{Introduction}
\label{sec:introduction}

Multimodal image matching is essential in various domains, including medical imaging, remote sensing, and autonomous driving. 
{
Its primary goal is to register images captured by different imaging sensors,
} thereby providing complementary   information for downstream tasks such as retinal disease monitoring~\cite{hernandez2021retinal}, laser eye  surgeries~\cite{truong2019glampoints}, image fusion~\cite{tian2023interpretable}, change detection~\cite{luo2023multiscale, li2023lightweight}, and image mosaicking~\cite{li2023real}. However, multimodal image matching faces significant challenges due to the inherent differences between modalities. A major challenge lies in the nonlinear intensity variations inherent between multimodal images. Additionally, geometric distortions and viewpoint discrepancies caused by different imaging conditions further complicate the matching process.


To address these challenges, many well-designed handcrafted and learning-based methods have been proposed. Early handcrafted approaches are generally divided into area-based~\cite{addison2015low,ye2017robust,ye2019fast} and feature-based methods~\cite{chen2010partial,fan2023robust,fan2024gls}. 
Area-based methods estimate image correspondences by calculating similarity metrics, such as mutual information~\cite{ritter1999registration} or cross-correlation~\cite{chanwimaluang2006hybrid}, between image patches. However, these methods often struggle with complex geometric transformations. 
Feature-based methods, such as SIFT~\cite{david2004distinctive} and its derivatives~\cite{dellinger2014sar, xiang2018sift}, mainly focus on achieving scale and rotation invariance by extracting distinctive keypoints and their associated descriptors, but they perform poorly with nonlinear intensity variations in multimodal images. 
Some approaches mitigate the impact of nonlinear intensity variations in multimodal images by introducing relative intensity relationships~\cite{chen2010partial,wang2015robust} between descriptors or by incorporating structured information from the images~\cite{li2019rift, fan2023robust}.

\begin{figure}[t]
    \centering
  \subfigure[Feature matching results]{\includegraphics[width=0.88\linewidth]{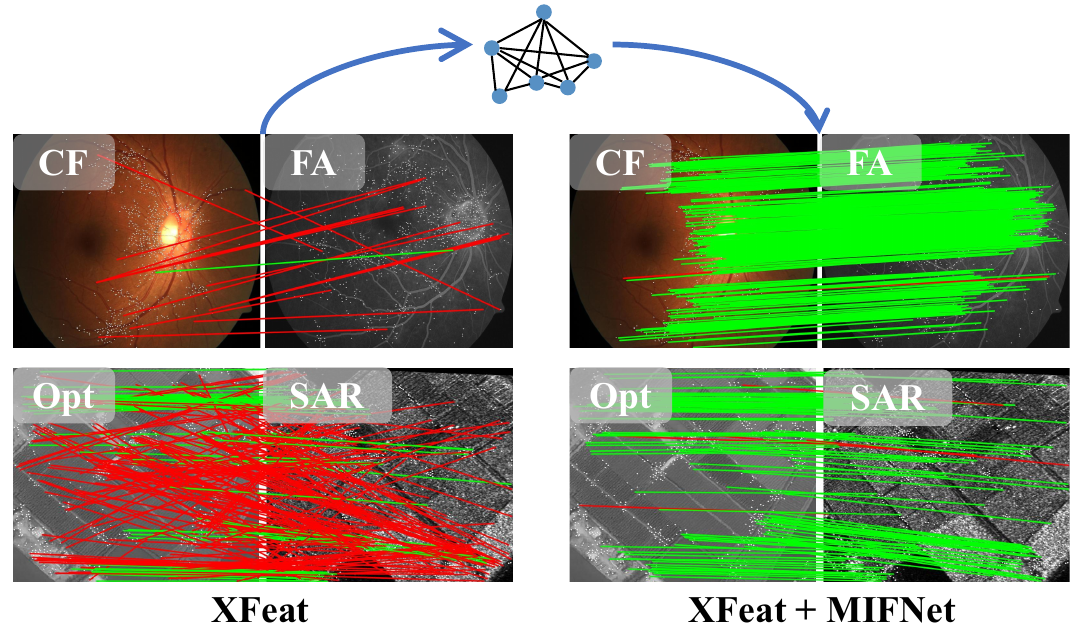}
  \label{fig:image1}}  
  \subfigure[Success registration rate  on different datasets]{\includegraphics[width=0.88\linewidth]{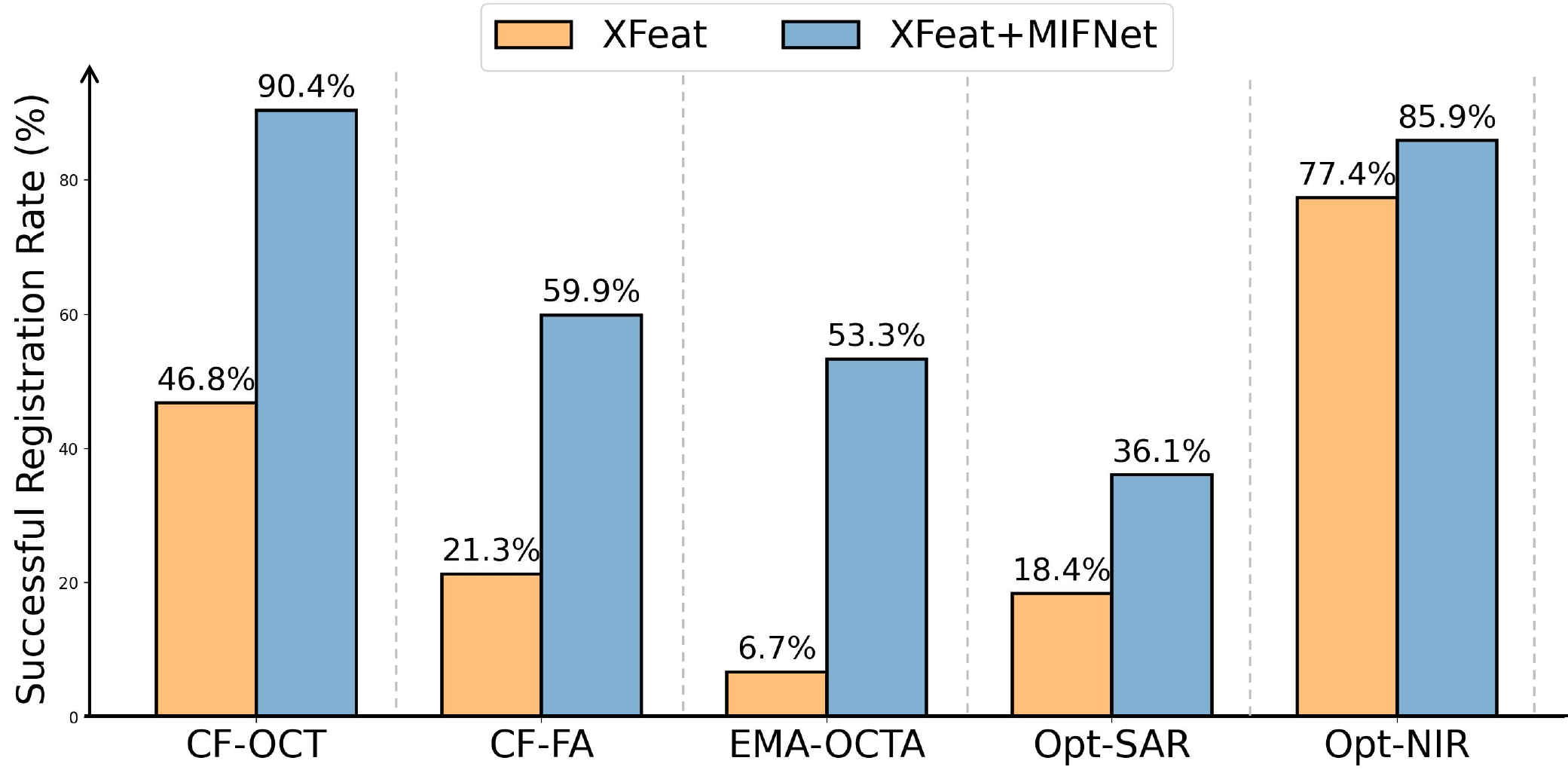}
  \label{fig:image2}}
    \caption{   
Cross modality matching performance of the recent single-modality detector XFeat~\cite{potje2024xfeat} enhanced by our MIFNet. (a) Green lines represent correct matches, while red lines represent incorrect matches. (b) The plot illustrates the performance improvement achieved by integrating MIFNet with the XFeat across five test datasets.
}
    \label{fig:mot_boost}
\end{figure}

In recent years, learning-based methods have achieved remarkable advancements in single-modality image matching by learning robust local feature representations through self-supervised training on single-modality data~\cite{detone2018superpoint, sarlin2020superglue, potje2024xfeat}.  However, extending these methods to multimodal image matching leads to a significant performance drop, as feature representations trained on single-modality datasets lack modality invariance, making them unreliable for matching across modalities.
To address this issue, some studies~\cite{deng2022redfeat, quan2022self} have explored learning modality-invariant features from paired cross-modal data. Nevertheless, these approaches rely on paired data, which is costly to obtain and thus presents significant challenges for widespread adoption. Other methods aim to bridge the gap between different modalities to simplify the matching by first constructing modality-specific style transfer\cite{sindel2022multi} or segmentation networks~\cite{zhang2021two}. However, the need for additional networks  for each modality limits the generalization of these methods to unseen modalities.

To address these challenges, we propose a novel approach for learning modality-invariant features from single-modality data, enabling them to serve as modality-agnostic keypoint descriptors for matching across unseen modalities. Our method builds on established learning-based frameworks for keypoint detection and description~\cite{detone2018superpoint, zhao2022alike, potje2024xfeat}, trained with single-modality data. The key challenge is learning modality-invariant features from such single-modality data. 
Inspired by the impressive generalization and representation capabilities of diffusion models~\cite{ho2020denoising, rombach2022high} across various applications~\cite{brempong2022denoising, saxena2023monocular, rahman2023ambiguous}, we explore the use of latent diffusion features to capture modality-invariant  keypoint description.
Specifically, we adopt the advanced graph neural network (GNN) techniques~\cite{sarlin2020superglue, lindenberger2023lightglue} to facilitate communication between keypoint descriptors (referred to as base features to distinguish them from subsequent features) and latent diffusion features through self-attention and cross-attention layers, trained in a self-supervised manner.
However, directly combining the latent diffusion features with the base features in this way results in suboptimal performance in multimodal matching.
In fact, although latent diffusion features exhibit some degree of cross-modal semantic invariance, they are coarse feature maps generated from image-text pairs, accompanied by noise. Direct fusion of latent diffusion features with the base features may lead to noise propagation, degrading the final feature quality.

To effectively utilize latent diffusion features, we first refine the coarse latent diffusion features before integrating them with the base features. Specifically, we introduce a novel Latent Feature Aggregation (LFA) module, which employs a Gaussian mixture model to refine coarse latent features, enhancing both semantic and modality invariance. 
{The choice of GMM is motivated by its ability to handle complex, overlapping feature distributions through soft clustering. This allows for better semantic separation and intra-cluster compactness compared to traditional regression models.}
{
Subsequently, we design a Cumulative Hybrid Aggregation (CHA) module to fuse the refined features with base features using multiple self-attention and cross-attention layers. The CHA module progressively aggregates features across multiple layers, facilitating both intra-image and cross-image feature interactions, thereby resulting in more precise feature representations.}
Both modules are trained in a self-supervised manner by comparing training samples with their randomly augmented counterparts, eliminating the need for additional annotations.
Remarkably, this process enables the learning of modality-invariant features using only single-modality training images. As shown in Fig.~\ref{fig:mot_boost}, MIFNet demonstrates strong modality invariance, significantly improving the cross-modality matching performance of state-of-the-art methods (\eg, XFeat~\cite{potje2024xfeat}) across diverse datasets. 

The main contributions of this work are as follows:
\begin{enumerate}
  \item We propose MIFNet, a novel framework for multimodal image matching that learns modality-invariant features from a keypoint descriptor trained on single-modality data, without needing data from other modalities.
  To the best of our knowledge, MIFNet is the first approach to learn modality-invariant features from single-modality data for multimodal image matching.
  
   \item MIFNet leverages pretrained diffusion models to enhance multimodal image matching. We introduce a novel latent feature aggregation module that refines coarse features using a Gaussian mixture model to improve their invariance. Additionally, a cumulative hybrid aggregation module efficiently combines base features and refined features. 

  \item  Extensive experiments on five different datasets demonstrate the effectiveness of our method and its strong zero-shot generalization. In addition, we also develop a challenging CF-FA    dataset for retinal image matching.
\end{enumerate}

\begin{figure*}[!t]
\centerline{\includegraphics[width=0.98\textwidth]{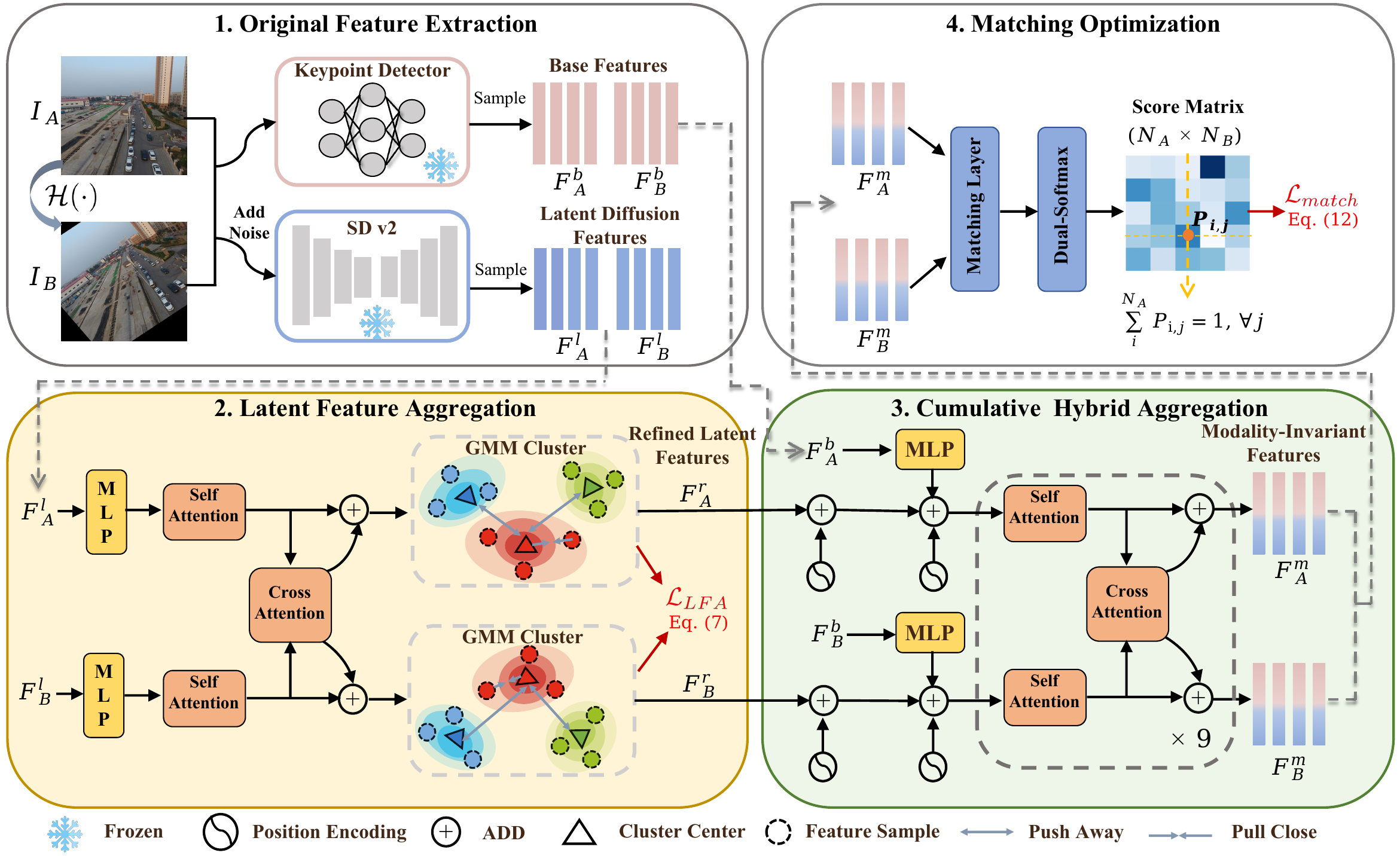}}
\caption{
Overview of our MIFNet framework. (1) Given an image $I_A$, we first apply a random transformation to obtain image $I_B$. Then, we extract base features ($F_{A}^{b}, F_{B}^{b}$) and latent diffusion features ($F_{A}^{l}, F_{B}^{l}$) using a frozen keypoint detector and the Stable Diffusion v2 model, respectively. (2) The proposed latent feature aggregation which includes self-attention, cross-attention and Gaussian Mixture Model (GMM) to refine the  aggregated latent features. (3) The proposed cumulative hybrid aggregation module integrates refined latent features and base features through multi-layer attention mechanisms to generate modality-invariant features (${F}_{A}^{m}, {F}_{B}^{m}$).  (4) Finally, the matching module predicts the matching score matrix between features and is supervised by the correspondence between images.
}
\label{fig:diff_frame}
\end{figure*}

\section{RELATED WORK}
\subsection{Single-modality Image Matching}
The core steps of single-modality  image matching include keypoint extraction, descriptor construction, and feature matching.  Handcrafted algorithms such as SIFT~\cite{david2004distinctive}, Harris corners~\cite{harris1988combined}, and ORB~\cite{rublee2011orb} are designed to deliver robust performance under challenging conditions, including variations in scale, rotation, illumination, and viewpoint.
With the advent of deep learning, numerous methods have been developed to leverage large-scale data for learning robust keypoint detection and description~\cite{detone2018superpoint, liu2022semi, shen2023detector, zhang2023perspectively, wang2024superjunction, potje2024xfeat}. 
SuperPoint~\cite{detone2018superpoint} employs a self-supervised approach to jointly train keypoint detection and description, demonstrating strong performance in homography estimation. SiLK~\cite{gleize2023silk} redefines each component from first principles, proposing a fully differentiable, lightweight, and flexible point detector.
Zhao \textit{et al.}~\cite{zhao2022alike} developed a partially differentiable keypoint detection module, achieving better performance with low inference latency. Building on~\cite{zhao2022alike}, the same authors proposed the ALIKED network\cite{zhao2023aliked}, which constructs deformable descriptors to enhance the geometric invariance of features.
More recently, XFeat~\cite{potje2024xfeat} redesigns the network architecture for keypoint detection and description and proposed an innovative matching refinement module, significantly boosting matching performance.
Additionally, methods like SuperGlue~\cite{sarlin2020superglue} and LightGlue~\cite{lindenberger2023lightglue} have achieved remarkable results by employing graph neural networks to aggregate positional information and descriptors through attention mechanisms. 
 
Although these methods have achieved strong results in single-modality scenarios, they face significant challenges in multimodal image matching due to modality differences. Our MIFNet effectively addresses this issue by incorporating latent diffusion features with the proposed LFA and CHA modules.

\subsection{Multimodal Image Matching}


The main challenges in multimodal image matching include geometric transformations and nonlinear intensity differences. To address these challenges, various well-designed handcrafted algorithms~\cite{addison2015low, zhang2018automated, ye2019fast, fan2021exploiting} and learning-based methods~\cite{lee2019deep, wang2021robust, deng2022redfeat,liu2024grid} have been proposed.

Early handcrafted methods perform matching by establishing a template window and calculating metrics such as normalized cross correlation ~\cite{ma2010fully} or mutual information~\cite{cole2003multiresolution} between image windows.
To address complex geometric transformations and  nonlinear radiometric variations, numerous methods have focused on constructing robust cross-modal features based on local keypoints.
Li \textit{et al.}~\cite{li2019rift} proposed to leverage phase congruence to detect  keypoints and introduce a maximum index map derived from a sequence of log-Gabor convolutions for feature description.
Fan \textit{et al.}~\cite{fan2023robust} developed an oriented filter-based matching approach, which specifically tackles the challenges posed by nonlinear intensity variations. However, when handling complex geometric transformations and cross-modal differences, handcrafted methods face performance limitations and high computational latency.

Learning-based multimodal image matching methods can be broadly categorized into data-driven and modality transformation-based approaches. 
Data-driven methods leverage paired datasets to train networks to learn modality-invariant features, as seen in works like ReDFeat~\cite{deng2022redfeat} and SDNet~\cite{quan2022self}. However, acquiring paired datasets is costly, and ground-truth pixel correspondences are difficult to obtain.
Modality transformation methods~\cite{ding2020weakly, wang2021robust, zhang2021two} aim to map different modalities to a unified representation before feature matching, reducing modality differences. 
Some studies~\cite{wang2021robust, zhang2021two} apply two segmentation networks to obtain binary vessel masks from multimodal fundus images before performing matching. These methods rely heavily on prior vascular information and require training separate segmentation networks for each modality, adding complexity.
Similarly, KPVSA-NET~\cite{sindel2022multi}, which leverages CycleGAN~\cite{zhu2017unpaired} for modality transfer, suffers from limited generalizability to unseen modalities due to its dependence on additional networks.

Different from these methods,   MIFNet requires only single-modality images as the training data to train the models for modality-invariant features, significantly reducing the cost to collect well {registered} multimodal data. Moreover, MIFNet demonstrates excellent generalization capabilities, achieving strong multimodal image matching performances across various unseen modality datasets.

\subsection{Latent Diffusion Features for Downstream Tasks}
Diffusion models~\cite{ho2020denoising,nichol2021improved,karras2022elucidating} decompose the image generation process into a sequence of steps using denoising autoencoders, achieving remarkable success in image synthesis. 
These models conceptualize image generation as learning the reverse of a known forward process. In particular, Stable Diffusion~\cite{rombach2022high} employs latent diffusion models (LDMs) that operate in the latent space of pretrained autoencoders, enabling the generation of high-resolution images. The integration of cross-attention layers in the model architecture further enhances LDMs, allowing them to produce detailed and high-quality images across a range of tasks.
Recently, a growing body of work has explored leveraging the powerful semantic representation capabilities of diffusion models for various downstream tasks~\cite{brempong2022denoising, saxena2023monocular, rahman2023ambiguous, tian2024diffuse}. For instance, Emmanuel \textit{et al.}\cite{brempong2022denoising} proposed a denoising-based decoder pre-training approach, yielding promising results in semantic segmentation. Similarly, Saxena \textit{et al.}\cite{saxena2023monocular} utilized diffusion features to predict dense depth from a single image.


In this paper, we investigate the challenge to use latent diffusion features for multimodal image matching and propose a novel self-supervised framework MIFNet to effectively aggregate base and latent diffusion features for modality-invariant features. 


\section{Method}

We employ a self-supervised training strategy to learn modality-invariant features from single-modality data. The overall framework of our approach is illustrated in Fig.~\ref{fig:diff_frame}. For each input image \( I_A \in \mathbb{R}^{W \times H \times 3} \), a paired image \( I_B \in \mathbb{R}^{W \times H \times 3} \) is generated using a random homographic transformation \( \mathcal{H}(\cdot) \). 
During the original feature extraction phase (Sec.~\ref{sec.m1}), we extract keypoints along with their corresponding base features and latent diffusion features. As discussed in the introduction, directly fusing the base and latent diffusion features can result in noise propagation, negatively affecting the final feature representation. To address this, we propose a Latent Feature Aggregation (LFA) module (Sec.~\ref{sec.m2}) to refine the latent features. 
Subsequently, we introduce a Cumulative Hybrid Aggregation (CHA) module (Sec.~\ref{sec.m3}), which facilitates interaction between the refined latent diffusion features and the base features, ultimately producing robust modality-invariant features for matching. Finally, the matching score matrix is supervised based on the established correspondence relationships (Sec.~\ref{sec.m4}).

\subsection{Original Feature Extraction}
\label{sec.m1}
 \medskip \noindent  \textbf{Base Feature Extraction.} 
Our method is a general approach that can be seamlessly integrated with various existing keypoint detection and description methods. In this paper, we use three different generic methods as baselines: SuperPoint \cite{detone2018superpoint}, ALIKED \cite{zhao2023aliked}, and XFeat \cite{potje2024xfeat}, along with a domain-specific algorithm, SuperRetina \cite{sindel2022multi}, for retinal image matching.
For each input image \( I_A \in \mathbb{R}^{W \times H \times 3} \), we generate a paired image \( I_B = \mathcal{H}(I_A) \in \mathbb{R}^{W \times H \times 3} \) using a random homographic transformation \( \mathcal{H}(\cdot) \).
Next, we extract keypoints along with their corresponding feature descriptors \( F_{A}^{b} \) and \( F_{B}^{b} \) for both \( I_A \) and \( I_B \), using the selected keypoint detection and description method. Here, \( F_{A}^{b} \) and \( F_{B}^{b} \) are referred to as the base features to distinguish them from features obtained in later stages.

\medskip \noindent  \textbf{Latent Diffusion Feature Extraction.} 
With advancements in diffusion and foundational models~\cite{nichol2021improved, rombach2022high, oquab2023dinov2}, various methods have been developed to extract features with strong cross-domain generalization capabilities. In this paper, we select Stable Diffusion v2~\cite{rombach2022high} (SDv2) as our feature extractor {thanks to its ability to encode rich and transferable semantic information, learned from billions of image-text pairs. Such pretraining enables the model to produce modality-invariant features even without task-specific fine-tuning}. 
The diffusion process in this model comprises forward and reverse stages. In the forward diffusion stage, noise is progressively added to clean images, transforming them from a coherent state to a noisy one. In contrast, the reverse diffusion process reconstructs the original image from the noisy state. 
To retain the model’s intrinsic capabilities, we extract features directly from the frozen decoder layer of the SDv2 during the forward diffusion process. {This design allows us to utilize the pretrained latent space as a plug-and-play semantic prior, preserving both high-level abstraction and spatial structure.} Specifically, we use intermediate features from the second layer of the decoder as the latent feature map. 
Finally, the latent diffusion features \( F_{A}^{l} \) and \( F_{B}^{l} \) are sampled from this latent feature map using the normalized coordinates of keypoints extracted by the base feature detector. 
{These latent features serve as high-level semantic cues that are less sensitive to low-level appearance variations, thereby enhancing the robustness and discriminability of our framework in challenging cross-modal matching scenarios.}

\subsection{ Latent Feature Aggregation}
\label{sec.m2}
The latent diffusion features derived from Stable Diffusion v2 exhibit strong semantic representation capabilities. However, as coarse feature maps, they contain noise that affects their semantic invariance. To address this, our module employs attention mechanisms to capture global contextual information and refines the semantic distribution of the aggregated features, enhancing the invariance of their semantic representations.

Specifically, we first reduce the dimensionality of the latent feature to 
$C$ using Multi-Layer Perceptron (MLP), and then apply self-attention and cross-attention to aggregate features both within each image and between different images to get the    refined features \( F_{A}^{r} \) and \( F_{B}^{r}\).

Taking feature aggregation on feature set $F_{A}^l$ as an example, the $i$-th feature $f_{A}^i$ is updated as follows:
\begin{equation}
f_{A}^i \leftarrow f_{A}^i + \text{MLP} \left( [f_{A}^i \mid M_{A \leftarrow O} \right] ),
\label{eq:self attention}
\end{equation}
where the symbol $\leftarrow$ represents the update operation, and 
$[\cdot \mid \cdot ]$ is the channel-wise concatenation.
The attention update message  $M_{A \leftarrow O}$ is computed as:
\begin{equation}
M_{A \leftarrow O} = \operatorname{Softmax}\left( \frac{W^q f_A^{i} (W^k f_O)^{\top}}{\sqrt{C}} \right) W^v f_O,
\label{eq:attention message}
\end{equation}
where $O$ corresponds to either $I_A$ or $I_B$, and $f_O$ represents the features of all keypoints from the respective image. In self-attention, $f_O$ comes from the same image as $f_A^{i}$, while in cross-attention, it is derived from the other image. $W^q$, $W^k$, and $W^v$ are linear projection matrices.

We propose to employ a separate Gaussian Mixture Model (GMM) for each image to cluster the aggregated latent features. 
The optimization objective is to improve the semantic distribution of latent features by enhancing the compactness of features within the same semantic region and increasing the separability between different semantic regions.
GMM is a probabilistic model that assumes all data points are generated from a mixture of Gaussian distributions, each characterized by its own mean and covariance. 
The GMM model clusters these features into $K$ classes. The probability density function of a GMM is defined as a weighted sum of $K$ Gaussian distributions:
\begin{equation}
    p(\mathbf{x}) = \sum_{k=1}^{K} \pi_k \mathcal{N}(\mathbf{x} \mid \boldsymbol{\mu}_k, \boldsymbol{\Sigma}_k),
\end{equation}
where $\pi_k$ is the mixture coefficient for the $k$-th Gaussian component.
$\mathcal{N}(\mathbf{x} \mid \boldsymbol{\mu}_k, \boldsymbol{\Sigma}_k)$ is a Gaussian distribution with mean $\boldsymbol{\mu}_k$ and covariance matrix $\boldsymbol{\Sigma}_k$.



{The GMM model parameters are optimized using the Expectation-Maximization (EM) algorithm, which iteratively refines cluster assignments and updates Gaussian parameters to maximize the likelihood of the observed feature distribution.}

To make the distance between features within the same semantic region more compact, we minimize the distance between features of the same class and their respective class center. The intra-class compactness loss is calculated as :
\begin{equation}
    \mathcal{L}_{\text{intra}} = \sum_{k=1}^{K} \sum_{\mathbf{x}_i \in \mathcal{C}_k} \|\mathbf{x}_i - \boldsymbol{\mu}_k\|_2^2,
\end{equation}
where $\mathcal{C}_k$ denotes the set of features belonging to the $k$-th class.

On the other hand, we aim to maximize the distance between classes, which allows the semantic features to have better separability. The inter-class loss is calculated as:
\begin{equation}
    \mathcal{L}_{\text{inter}} =  \sum_{k=1}^{K} \sum_{j \neq k} \|\boldsymbol{\mu}_k - \boldsymbol{\mu}_j\|_2^2.
\end{equation}

{
We employ the combination of the intra-class compactness and inter-class separation objectives to refine the semantic distribution of the latent features.} 

\subsection{Cumulative  Hybrid Aggregation}
\label{sec.m3}

In this module, we first enable interaction between the refined diffusion features \( F_{A}^{r} \) and \( F_{B}^{r} \) and the base features \( F_{A}^{b} \) and \( F_{B}^{b} \), which contain strong geometric details that enhance the matching performance by providing more precise geometric information. This interaction is facilitated through a multi-layer aggregation module, which progressively accumulates features to produce modality-invariant and robust representations.

Specifically, we use MLP layers to standardize the dimensionality of the base features to a common dimension \( C \). Then, positional encoding (PE) is applied to incorporate keypoint location information \( P_{loc} \). We use an MLP layer to encode the original keypoint vector into \( \mathbb{R}^{N \times C} \), and the encoded positional information is then added to the refined diffusion and base features, which are fused to obtain the initial modality-invariant features \( {F}^m \).

\begin{equation}
{F}^m =(\text{PE}(P_{loc}) + {F}^{r}) + (\text{PE}(P_{loc}) + \text{MLP}({F}^{b})).
\label{eq:feature align}
\end{equation}

Subsequently, the  features ${F}^m$ are iteratively aggregated and updated through multiple layers. In each layer, we alternately apply self-attention and cross-attention units. 
The calculation process of the mixed features is described as follows:
\begin{equation}
f_{j+1}^m \leftarrow f_{j}^m  + \text{MLP} \left( [f_{j}^m  \mid M \right]), 
\label{eq:multi feature update}
\end{equation}
where \( j \) represents the layer index, and \(f_{j}^m\) belongs to \( F_{A,j}^m \) or \( F_{B,j}^m \). In our experiments, we employ a total of 9 layers of attention-based aggregation. The calculation of \( {M} \) is defined in Eq.~\eqref{eq:attention message}. 

Through progressive feature aggregation, we gradually allow the diffusion feature to interact with the distinctive base feature, enabling the output features to obtain modality-invariance ability.

\begin{figure*}[t]
\centerline{\includegraphics[width=0.98\textwidth]{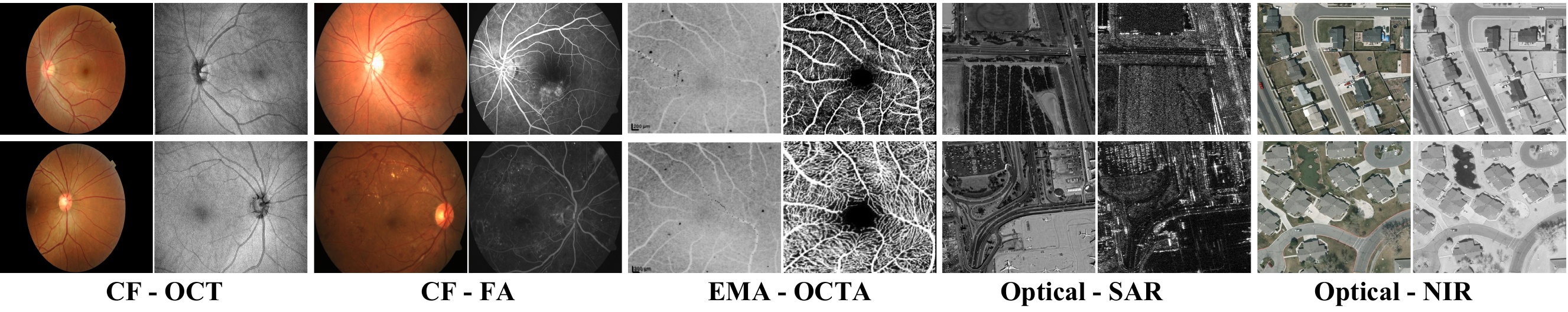}}
\caption{Some examples of involved multimodal image matching test datasets.
}
\label{fig:image_introduce}
\end{figure*}

\subsection{Matching Optimization}
\label{sec.m4}

During the training phase, we apply random homography transformations $\mathcal{H}(\cdot)$ to generate pairs of single-modality color images as input. In the feature extraction stage, we first obtain the keypoint  from the image pairs. 
Subsequently, we calculate the reprojection error of keypoints based on the $\mathcal{H}(\cdot)$, and obtain the ground-truth matching relationships between keypoints.

We apply point-level matching supervision to the output features of each layer in the cumulative hybrid aggregation stage. Specifically, for the output features $F_A^m$ and $F_B^m$ of each layer, we use a matching layer to compute a matching score matrix $S$:
\begin{equation}
S = F_A^m \cdot (F_B^m)^\top \in \mathbb{R}^{N_A \times N_B},
\label{eq:score cal}
\end{equation}
where \( N_A \) and \( N_B \) represent the number of keypoints in the image pair $I_A$ and $I_B$, respectively. We then apply a dual softmax operation to compute the predicted matching probability $P_{i,j}$ for each keypoint pair. The dual softmax operation ensures that all predicted probabilities remain less than 1.
In addition, some keypoints may have no matching pairs. Following ~\cite{lindenberger2023lightglue}, we introduce a learnable parameter for each sample to predict the probability of a match existing for each keypoint:
\begin{equation}
\sigma_i = \text{Sigmoid} (\text{Linear}(f_i)).
\label{eq:predict unmatch}
\end{equation}

We minimize the log-likelihood of the predicted assignments for each layer and compute the average matching loss across all layers:
\begin{equation}
\begin{split}
\mathcal{L}_{match} = &-\frac{1}{|M_{gt}|} \sum_{(i,j) \in M_{gt}} \log P_{i,j} \\
         &+ \frac{1}{|U_A|} \sum_{i \in U_A} \log \sigma_i^{U_A} 
         + \frac{1}{|U_B|} \sum_{j \in U_B} \log \sigma_j^{U_B},
\label{eq:match loss}
\end{split}
\end{equation}
where $M_{gt}$ denotes the set of the matched points, and \( U_A \) and \( U_B \) denote the sets of unmatched points in $I_A$ and $I_B$, respectively.


{
In addition to the matching supervision, we introduce a latent feature aggregation loss to enhance the semantic consistency of the extracted features. The LFA loss is defined as:
\begin{equation}
    \mathcal{L}_{\text{LFA}} = \mathcal{L}_{\text{intra}} - \mathcal{L}_{\text{inter}},
\end{equation}
where $\mathcal{L}_{\text{intra}}$ encourages features within the same semantic cluster to be compact, and $\mathcal{L}_{\text{inter}}$ enforces separation between different cluster centers.
}
The final training objective combines the matching loss and the LFA loss as:
\begin{equation}
    \mathcal{L} = \mathcal{L}_{\text{match}} + \lambda \mathcal{L}_{\text{LFA}},
\end{equation}
where $\lambda$ is a weighting factor set to 2.0 in all experiments unless otherwise specified.

\section{Experiments}

\renewcommand{\arraystretch}{1.15} 

\begin{table*}[ht]

\centering 
\caption{Zero-shot matching performance on multimodal retinal image datasets, including CF-OCT, CF-FA and EMA-OCTA. T: Traditional methods, which do not require training data; M: Methods that require multimodal training data; S: Methods that only require color fundus image training data. The best results are highlighted in bold.
}
\label{tab:regis_result}
\resizebox{1.0\linewidth}{!}{
\begin{tabular}{l|c|ccc|ccc|ccc}
\hline
\multicolumn{1}{l|}{\multirow{2}{*}{Methods}} & \multirow{2}{*}{Type}& \multicolumn{3}{c|}{CF - OCT}    & \multicolumn{3}{c|}{CF - FA} & \multicolumn{3}{c}{EMA - OCTA}   \\ 
\cline{3-11}
\multicolumn{1}{l|}{}  &    
& \multicolumn{1}{c}{${\text{SRR} (\%)}$ $\uparrow$} &\multicolumn{1}{c}{RMSE  $\downarrow$}  
& \multicolumn{1}{c|}{${\text{MAE}}$ $\downarrow$ } &
\multicolumn{1}{c}{${\text{SRR} (\%)}$ $\uparrow$} &\multicolumn{1}{c}{RMSE  $\downarrow$}  
& \multicolumn{1}{c|}{${\text{MAE}}$ $\downarrow$ }&
\multicolumn{1}{c}{${\text{SRR} (\%)}$ $\uparrow$} &\multicolumn{1}{c}{RMSE $\downarrow$}  
& \multicolumn{1}{c}{${\text{MAE}}$ $\downarrow$  } \\
\hline
\multicolumn{1}{l|}{Surf-PIFD-RPM\scriptsize\textit{~(BSPC'15)}}& \multirow{2}{*}{T}  & 24.5 & 43.9 & 91.2 & 34.9 & 56.2 & 115.6 & 6.7 & 48.4 & 113.2  \\
\multicolumn{1}{l|}{UR-SIFT-PIFD\scriptsize\textit{~(ICPR'18)}}&    & 27.4 & 40.1 & 87.6 & 29.9 & 63.4 & 140.4 & 6.7 & 56.9 & 102.7  \\
\hline
\multicolumn{1}{l|}{KPVSA-NET\scriptsize\textit{~(MICCAI'22)}}& \multirow{3}{*}{M} & 63.9 & 14.3 & 37.5 & 50.4 & 34.6 & 81.9 & 40.0 & 22.4 & 47.1  \\
\multicolumn{1}{l|}{Content-Adaptive\scriptsize\textit{~(TIP'21)}} &  & 75.7 & 8.9 & 24.7 & 46.9 & 40.3 & 85.3 & 46.7 & 17.9 & 38.6  \\
\multicolumn{1}{l|}{{Xoftr\scriptsize\textit{~(CVPR'24)}}} &  & 59.6 & 17.8 & 38.5 & 52.0 & 29.2 & 64.3 & 33.3 & 20.7 & 42.3  \\
\hline
\multicolumn{1}{l|}{{Superglue\scriptsize\textit{~(CVPR'20)}}} &\multirow{11}{*}{S} & 76.9  & 7.9  & 16.1 & 20.8  & 58.3 & 127.6 & 6.7 & 37.5 & 64.3\\
\multicolumn{1}{l|}{{lightglue\scriptsize\textit{~(ICCV'23)}}} & & 78.8  & 7.3  & 14.6 & 24.3  & 52.7 & 101.1 & 6.7 & 33.1 & 56.5\\
\multicolumn{1}{l|}{{GIM (DKM)\scriptsize\textit{~(ICLR'24)}}} & & 80.8  & 5.3  & 12.7 & 41.8  & 39.4 & 90.3 & 6.7 & 32.1 & 59.8\\
\multicolumn{1}{l|}{SuperPoint\scriptsize\textit{~(CVPR'18)}} & & 73.1  & 9.7 & 29.2 & 17.4  & 62.0 & 145.1 & 6.7 & 41.3 & 85.1\\
\multicolumn{1}{l|}{SuperPoint + MIFNet} &  & \cellcolor{gray!10} 88.6\redup{15.5}& \cellcolor{gray!10} 4.2\bluedown{5.5} & \cellcolor{gray!10} 10.9\bluedown{18.3} & \cellcolor{gray!10} 64.1\redup{46.7}& \cellcolor{gray!10} 17.7\bluedown{44.3}&\cellcolor{gray!10} {49.5\bluedown{95.6}} &\cellcolor{gray!10} 20.0\redup{13.3} &\cellcolor{gray!10} 29.9\bluedown{11.4} &\cellcolor{gray!10} 62.5\bluedown{22.6}\\

\multicolumn{1}{l|}{SuperRetina\scriptsize\textit{~(ECCV'22)}}&  & 67.3 & 7.7 & 22.3 & 29.5 & 63.3 & 144.8 & 26.7 & 28.9 & 58.7  \\
\multicolumn{1}{l|}{ SuperRetina + MIFNet}&  &\cellcolor{gray!10} 84.9\redup{17.6} &\cellcolor{gray!10} 4.4\bluedown{3.3} &\cellcolor{gray!10} 11.5\bluedown{10.8} &\cellcolor{gray!10} \textbf{66.0}\redup{36.5}&\cellcolor{gray!10} {14.6}\bluedown{48.7}&\cellcolor{gray!10} {45.0}\bluedown{99.8}&\cellcolor{gray!10} \textbf{73.3}\redup{46.6} &\cellcolor{gray!10} \textbf{8.3}\bluedown{20.6} &\cellcolor{gray!10} \textbf{17.5}\bluedown{41.2} \\

\multicolumn{1}{l|}{ALIKED\scriptsize\textit{~(TIM'23)}} &   & 39.8  & 20.1  & 51.3 & 10.3 & 87.2 & 169.1 & 13.3 & 39.4 & 71.2 \\
\multicolumn{1}{l|}{ALIKED + MIFNet}&  &\cellcolor{gray!10}  73.1\redup{33.3} &\cellcolor{gray!10}  7.5\bluedown{12.6} &\cellcolor{gray!10}  20.2\bluedown{31.1} &\cellcolor{gray!10}  50.6\redup{40.3} &\cellcolor{gray!10}  12.7\bluedown{74.5}  &\cellcolor{gray!10} 36.1\bluedown{133.0} &\cellcolor{gray!10} 46.7\redup{33.4} &\cellcolor{gray!10} 11.9\bluedown{27.5} &\cellcolor{gray!10} 26.1\bluedown{45.1}\\

\multicolumn{1}{l|}{XFeat\scriptsize\textit{~(CVPR'24)}} & & 46.8 & 16.2 & 43.6 & 21.3 & 70.7 & 149.4 & 6.7 & 41.5 &  71.9 \\
\multicolumn{1}{l|}{XFeat + MIFNet}&  &\cellcolor{gray!10} \textbf{90.4}\redup{43.6} &\cellcolor{gray!10} \textbf{3.9}\bluedown{12.3} &\cellcolor{gray!10} \textbf{10.1}\bluedown{33.5} &\cellcolor{gray!10} 59.9\redup{38.6}&\cellcolor{gray!10} \textbf{11.4}\bluedown{59.3}&\cellcolor{gray!10} \textbf{35.6}\bluedown{113.8}&\cellcolor{gray!10} 53.3\redup{46.6} &\cellcolor{gray!10} 10.2\bluedown{31.3} &\cellcolor{gray!10} 24.0\bluedown{47.9} \\

\hline
\end{tabular}
}
\end{table*}

\subsection{Datasets}
To comprehensively evaluate the zero-shot capability of our proposed algorithm, we conduct assessments in two distinct perspectives: retinal fundus images and remote sensing images.
For multimodal retinal image matching experiments, we use the MeDAL-Retina Dataset~\cite{nasser2024reverse} as the training set, which includes 1,900 unlabeled color fundus images.  Additionally, we collect three   cross-modal retinal image datasets for testing.
In the remote sensing scenario, we select 1,200 color images from each of the SEN1-2 dataset~\cite{schmitt2018sen1} and the VisDrone dataset~\cite{zhu2021detection} as the training data.
Additionally, we use  two publicly available aerial-view datasets for testing. 
The details of the test datasets are as follows:

\smallskip  \textbf{(1) The CF-OCT dataset} contains 52 pairs of color fundus and optical coherence tomography images~\cite{lee2015registration}.
The color fundus images range in size from $410 \times 410$ to $1016 \times 675$, and the OCT en-face images range from $304 \times 304$ to $513 \times 385$. Fundus images were captured with the TRC-NW8 camera, and OCT images with the Topcon DRI OCT-1 system. The main challenges include the field-of-view differences and speckle noise in OCT images. Each pair is annotated with 8 corresponding points.

\smallskip  \textbf{(2) The CF-FA dataset} contains color fundus and fluorescein angiography images collected locally from 200 anonymous patients with various retinal diseases including diabetic retinopathy, age-related macular degeneration, macular edema, retinal vein occlusion, etc. It includes 400 image pairs, with high-resolution images (up to $2588 \times 1958$) that exhibit different perspective changes due to various diseases. Each image pair is annotated with 10 corresponding points.


\smallskip  \textbf{(3) The EMA-OCTA dataset~\cite{wang2023memo}} contains erythrocyte-mediated angiography (EMA) and optical coherence tomography angiography (OCTA) images and is publicly available. It consists of superficial vascular plexus projections from OCTA scans, offering clear views of arteries and veins, which are also visible in the EMA images. The dataset exhibits variations in vessel thickness, quantity, and significant contrast differences. For testing, we use 15 image pairs, each annotated with 6 pairs of corresponding points.

\smallskip  \textbf{(4) The Optical-SAR dataset~\cite{xiang2020automatic}} contains SAR images captured by the Chinese GaoFen-3 (GF-3) satellite, equipped with a multipolarized C-band SAR sensor. Operating in spotlight mode, GF-3 provides geocoded images with 1-meter resolution and 10-km ground coverage. The paired images have a resolution of $512 \times 512$ pixels. For testing, we select 400 image pairs from the dataset.

\smallskip  \textbf{(5) The Optical-NIR Dataset}. We use the VEDAI dataset~\cite{razakarivony2016vehicle} as the Aerial Optical-NIR dataset. 
Due to weather variations and object scale changes, this dataset is often used to validate multimodal image matching methods. The image pairs have a resolution of 512×512 pixels. We randomly select 200 pairs images for testing.

Examples of some sampled images are shown in Fig.~\ref{fig:image_introduce}.

\subsection{Implementation Details}
During MIFNet training, {we randomly select a single image from the training set and apply various random augmentations, such as homography transformations and brightness adjustments, to generate a synthetic training pair consisting of the original and the augmented image.}
These pairs are separately processed through a pre-trained frozen keypoint detector and Stable Diffusion v2~\cite{rombach2022high} to extract keypoints, base features, and diffusion features. Using pixel correspondences between the image pairs, we supervise the training of MIFNet. 
{In the LFA module, the number of clusters $K$ for GMM is set to 5. We adopt the standard \texttt{GaussianMixture} implementation from \texttt{scikit-learn}, using K-means for initialization and the Expectation-Maximization (EM) algorithm for parameter optimization.}
We use the Adam optimizer with a learning rate of $1 \times 10^{-4}$, a batch size of 2, and 15 epochs for training. The model is trained on an NVIDIA 3090 GPU. For multimodal image matching evaluation, we use K-nearest neighbors for feature matching and apply \texttt{cv2.findHomography} to compute the homography matrix.

\subsection{Zero-shot Evaluation on Multimodal Retinal Image Matching}
\label{sec:result_retina}

\medskip \noindent \textbf{{Experimental Setting.}} 
To verify the effectiveness of our method, we train the MIFNet on single-modality color fundus images and test on three challenging datasets under a zero-shot setting without any fine-tuning. We use keypoint detection and description methods including SuperPoint~\cite{detone2018superpoint}, SuperRetina~\cite{liu2022semi}, ALIKED~\cite{zhao2023aliked}, and XFeat~\cite{potje2024xfeat} as base feature extractors combined with MIFNet to validate its effectiveness. The base detectors are fine-tuned on the color fundus training set before being frozen during MIFNet training.
We compare our methods ($X$+MIFNet, where $X$ denotes different base detectors) with five multimodal retinal image matching approaches, including traditional handcrafted methods SURF-PIIFD-RPM~\cite{wang2015robust} and UR-SIFT-PIIFD~\cite{zhang2018automated}, which do not require training, as well as KPVSA-NET~\cite{sindel2022multi}, Content-Adaptive~\cite{wang2021robust}, and {Xoftr~\cite{tuzcuouglu2024xoftr}}, which are optimized using domain-specific data. 
We make our best efforts to collect publicly available datasets for retraining. These include HRF~\cite{odstrvcilik2009improvement}, RECOVERY-FA~\cite{Ding_2020_TIP_DeepVesselSeg4FA}, ROSE~\cite{ma2020rose}, and MEMO~\cite{wang2023memo}.
{In addition, we also compare against several representative state-of-the-art single-modality matching methods, including SuperGlue~\cite{sarlin2020superglue}, LightGlue~\cite{lindenberger2023lightglue}, and GIM~\cite{xuelun2024gim}, to further validate the performance gap under cross-modal settings.}

\medskip \noindent \textbf{{Evaluation Metrics.}} Following the evaluation criteria established in~\cite{lee2019deep}, we employ three metrics to quantitatively assess performance: root mean square error ({RMSE}), maximum absolute error ({MAE}), and success registration rate (SRR). 

We begin by estimating the homography transformation between image pairs based on the predicted matched points. The keypoints from the query image are then mapped onto the reference image, and the RMSE and MAE are computed for each image pair using the ground truth keypoint annotations. If fewer than four keypoints are matched, the image pair is considered a failure and excluded from the RMSE and MAE calculations. In this experiment, the metrics are calculated on images of size $768 \times 768$. An image pair is considered successfully registered if its RMSE is less than 10 and MAE is less than 20. The Success Rate of Registration (SRR) is then defined as the proportion of successfully registered images.

\begin{figure*}[t]
\centerline{\includegraphics[width=0.98\textwidth]{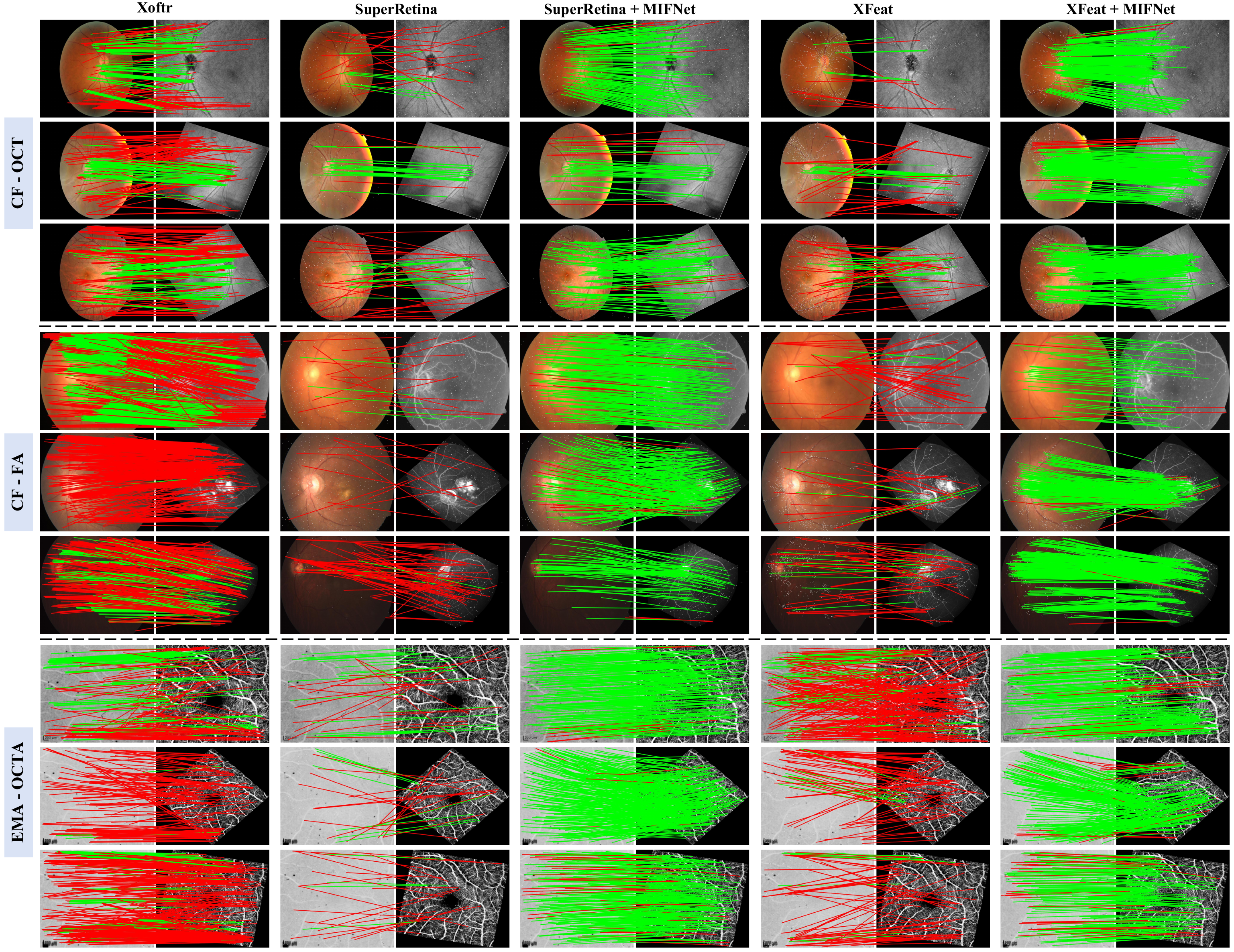}}
\caption{The visualization results on three multimodal retinal test sets CF-OCT, CF-FA, and EMA-OCTA. The green lines represent correctly matched pairs, while the red lines represent incorrectly matched pairs. {Each scene includes the results of tests under geometric transformations}.}
\label{fig:vis_result}
\end{figure*}

\renewcommand{\arraystretch}{1.15} 

\begin{table*}[!t]
\centering
\caption{Zero-shot matching performance on multimodal remote sensing image matching datasets, including Optical-SAR and Optical-NIR. T: Traditional methods, which do not require training data; M: Methods that require multimodal training data; S: Methods that only require single-modality optical image training data. The best results are highlighted in bold.}
\label{tab:remote_result}
\resizebox{0.8\linewidth}{!} {
\begin{tabular}{l|c|ccc|ccc}
\hline
\multicolumn{1}{l|}{\multirow{2}{*}{Methods}}& \multirow{2}{*}{Type} & \multicolumn{3}{c|}{Optical - SAR}   & \multicolumn{3}{c}{Optical - NIR}    \\ 
\cline{3-8}
\multicolumn{1}{l|}{}  &   
& \multicolumn{1}{c}{${\text{SRR} (\%)}$ $\uparrow$} &\multicolumn{1}{c}{$H_{\text{err}}$  $\downarrow$}  
& \multicolumn{1}{c|}{MS (\%) $\uparrow$ } &
\multicolumn{1}{c}{${\text{SRR} (\%)}$ $\uparrow$} &\multicolumn{1}{c}{$H_{\text{err}}$  $\downarrow$}  
& \multicolumn{1}{c}{MS (\%) $\uparrow$ }\\
\hline 
\multicolumn{1}{l|}{OS-SIFT\scriptsize\textit{~(TGRS'18)}} &\multirow{2}{*}{T} 
& 18.6 & 4.0 & 3.1 & 77.7 & 1.7 & 34.1  \\
\multicolumn{1}{l|}{RIFT2\scriptsize\textit{~(Arxiv'23)}} & & 27.5 & 4.0 & 4.9 & 79.8 & 1.6 & 39.2  \\
\hline
\multicolumn{1}{l|}{CMMNet\scriptsize\textit{~(TGRS'21)}} &\multirow{3}{*}{M} & 22.8  & 4.3 & 2.6 & 80.6 & 1.5 & 39.4  \\
\multicolumn{1}{l|}{ReDFeat\scriptsize\textit{~(TIP'23)}} & & \textbf{41.4} & \textbf{3.3} & \textbf{9.1} & 85.4 & 1.5 &  46.8  \\
\multicolumn{1}{l|}{{Xoftr\scriptsize\textit{~(CVPR'24)}}} & & {32.8} & {3.8} & {7.7} & 86.0 & 1.4 &  44.9  \\
\hline
\multicolumn{1}{l|}{{Superglue\scriptsize\textit{~(CVPR'20)}}} &\multirow{9}{*}{S} &18.0  & 4.3  & 3.0 & 79.0 & 1.8 & 38.9   \\
\multicolumn{1}{l|}{{Lightglue\scriptsize\textit{~(ICCV'23)}}} & &19.3  & 4.3  & 3.2 & 81.5 & 1.7 & 39.4   \\
\multicolumn{1}{l|}{{GIM (DKM)\scriptsize\textit{~(ICLR'24)}}} & &20.5  & 4.2  & 3.8 & 82.5 & 1.6 & 42.0   \\
\multicolumn{1}{l|}{SuperPoint\scriptsize\textit{~(CVPR'18)}} & & 15.1  & 4.4  & 2.8 & 74.9 & 1.9 & 35.6   \\
\multicolumn{1}{l|}{SuperPoint + MIFNet} & &\cellcolor{gray!10} 39.7\redup{24.6} &\cellcolor{gray!10} 3.4\bluedown{1.0} &\cellcolor{gray!10} 8.5\redup{5.7} &\cellcolor{gray!10} 84.8\redup{9.9} &\cellcolor{gray!10} 1.5\bluedown{0.4} &\cellcolor{gray!10} \textbf{51.1}\redup{15.5}\\
\multicolumn{1}{l|}{ALIKED\scriptsize\textit{~(TIM'23)}} & & 8.3  & 4.5  & 1.9 & 79.6 & 1.8 & 41.1  \\

\multicolumn{1}{l|}{ALIKED + MIFNet} & &\cellcolor{gray!10} 30.2\redup{21.9} &\cellcolor{gray!10} 3.9\bluedown{0.6} &\cellcolor{gray!10} 5.6\redup{3.7} &\cellcolor{gray!10} \textbf{88.6}\redup{9.0} &\cellcolor{gray!10} \textbf{1.3}\bluedown{0.5} &\cellcolor{gray!10} {48.9}\redup{7.8}\\
\multicolumn{1}{l|}{XFeat\scriptsize\textit{~(CVPR'24)}} & & 18.4  & 4.3  & 3.0 & 77.4 & 1.9 & 37.8  \\

\multicolumn{1}{l|}{XFeat + MIFNet} & &\cellcolor{gray!10} 36.1\redup{17.7} &\cellcolor{gray!10} 3.6\bluedown{0.7} &\cellcolor{gray!10} 8.1\redup{5.1} &\cellcolor{gray!10} 85.9\redup{8.5} &\cellcolor{gray!10} 1.5\bluedown{0.4} &\cellcolor{gray!10} 46.1\redup{8.3} \\
\hline
\end{tabular}
}
\end{table*}

\medskip \noindent \textbf{{Quantitative Results.}}
Tab.~\ref{tab:regis_result} shows the quantitative registration results of different datasets.  
We compare the results of handcrafted-based methods, learning-based multimodal methods and single-modality methods.

On the CF-OCT test set, learning-based single-modality matching algorithms outperform all handcrafted methods~\cite{wang2015robust,zhang2018automated}, with
{
GIM~\cite{xuelun2024gim}
achieving the highest registration success rate of 80.8\%.} This is due to the primary challenge in the CF-OCT dataset being the scale difference between image pairs, which learning-based methods handle more effectively.
For multimodal-specific methods, Content-Adaptive~\cite{wang2021robust} achieves an accuracy of 75.7\%, outperforming SuperPoint by converting images into binary vessel masks using a vessel segmentation network. In contrast, KPVSA-NET~\cite{sindel2022multi} performs poorly, as its accuracy is constrained by the effectiveness of the CycleGAN network.
However, after incorporating our proposed MIFNet, the SuperPoint, SuperRetina, and XFeat methods surpass existing multimodal-specific methods. Notably, MIFNet leads to a 43.6\% improvement in the success rate when applied to XFeat~\cite{potje2024xfeat}, achieving the highest success rate of 90.4\%.

The CF-FA dataset exhibits both intensity and geometric differences between modalities, leading to a significant performance drop for single-modality methods. In particular, the reversal of vessel and background colors between FA and CF image pairs presents a considerable challenge for local feature matching. Traditional handcrafted algorithms struggle due to the nonlinear intensity differences across modalities. Multimodal methods achieve a success rate of 50.4\% by mitigating these modality differences. However, after incorporating MIFNet, SuperPoint, SuperRetina, ALIKED , and XFeat show improvements in success rates of 46.7\%, 36.5\%, 40.3\%, and 38.6\%, respectively. SuperRetina~\cite{liu2022semi} achieves the highest success rate of 66.0\%, surpassing KPVSA-NET~\cite{sindel2022multi} at 50.4\%. This demonstrates that enhancing the cross-modal robustness of single-modality algorithms can significantly improve overall matching performance.

The EMA-OCTA dataset exhibits the most significant modality differences, particularly in vessel thickness. OCTA data captures finer and smaller vessels, while EMA data shows low contrast, further complicating the matching process. We observe that existing methods perform poorly under these conditions. However, feature fusion with MIFNet significantly improves the accuracy of the three single-modality baselines. Notably, MIFNet boosts the success rate by 46.6\% when applied to SuperRetina and achieving the highest success rate of 73.3\%. For the baselines ALIKED and XFeat, which perform poorly, MIFNet still brings improvements of 33.4\% and 46.6\% in SRR metric, respectively. This demonstrates the generalizability of our method.

Fig.~\ref{fig:vis_result} shows several feature matching results for visualization. As we can see, MIFNet achieves more accurate keypoint matching, leading to improved registration performance. 

\begin{figure*}[!t]
\centerline{\includegraphics[width=0.98\textwidth]{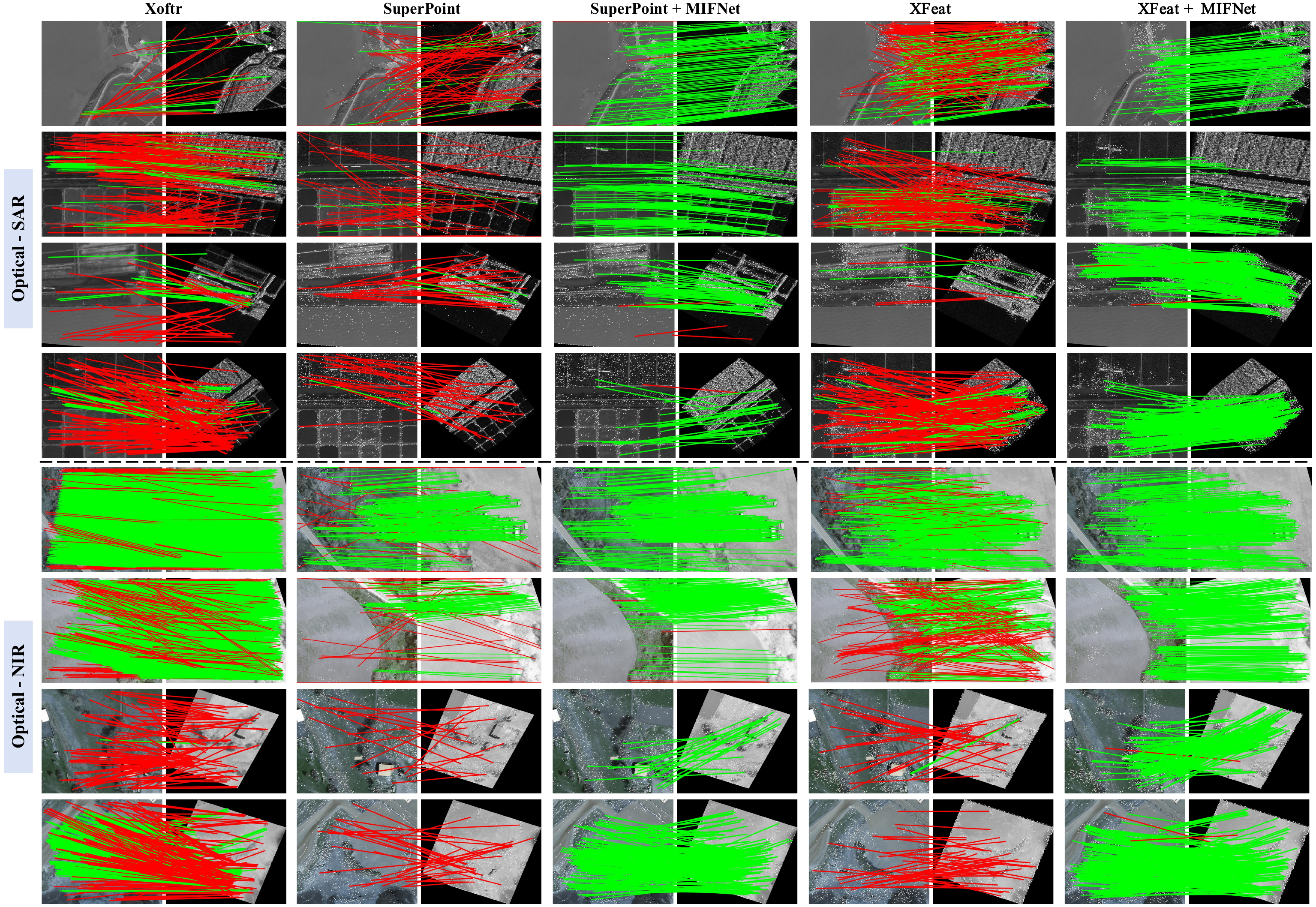}}
\caption{The visualization results of feature matching in aerial scenes: Optical-SAR and Optical-NIR. The green lines represent correctly matched pairs according to the homography ground truth, while the red lines represent incorrectly matched pairs. {Each scene includes the results of tests under geometric transformations}.
}
\label{fig:vis_match_remote}
\end{figure*}

\subsection{Zero-shot Evaluation  on Multimodal Remote Sensing Image Matching}
\label{sec:result_remote}

\medskip \noindent \textbf{{Experimental Setting.}} 
Our method is trained on a single-modality optical dataset and directly tested on two unseen datasets without any fine-tuning. We apply MIFNet to three representative single-modality methods: SuperPoint~\cite{sarlin2020superglue}, ALIKED~\cite{zhao2023aliked}, and XFeat~\cite{potje2024xfeat}. 
{In addition, we also compare our method with several state-of-the-art single-modality matchers, including SuperGlue~\cite{sarlin2020superglue}, LightGlue~\cite{lindenberger2023lightglue}, and GIM~\cite{xuelun2024gim}.}
We then compare our approach with two traditional methods, RIFT2~\cite{li2023rift2} and OS-SIFT~\cite{xiang2018sift}, which are widely used in multimodal remote sensing image matching, as well as two multimodal methods, CMMNet~\cite{cui2021cross}, ReDFeat~\cite{deng2022redfeat} and {Xoftr~\cite{tuzcuouglu2024xoftr}}. Notably, these methods require paired data specific to each modality pair for training. To facilitate this, we select 2,000 image pairs from the GF3~\cite{xiang2020automatic} dataset and 1,000 image pairs from the VEDAI~\cite{razakarivony2016vehicle} dataset for their training.

\medskip \noindent \textbf{{Evaluation Metrics.}}
The remote sensing test set consists of paired images. During testing, we construct test pairs using random transformations, including random rotations within the range of [-20°, 20°], distortion scale within [0, 0.2], and random scaling within [0.8, 1.2]. To evaluate the performance, we utilize the homography matrix error {$H_{\text{err}}$}, successful registration rate {(SRR)} and matching score {(MS)} as metrics~\cite{deng2022redfeat}.

Following ReDFeat~\cite{deng2022redfeat}, we first compute the error between the predicted homography matrix $\hat{H}$ and the ground-truth homography matrix  $H$, as defined by the following formula:
\begin{equation}
H_{\text{err}} = \lVert H - \hat{H} \rVert_2.
\end{equation}
In the computation of SRR here, a pair of images is considered successfully registered if $H_{\text{err}}$ is less than a threshold. We set the threshold to 5 in our experiments.
Matching Score (MS) is another commonly used metric~\cite{detone2018superpoint, deng2022redfeat}, which is calculated as the ratio of the number of correctly matched point pairs within the overlap region to the total number of detected keypoints.

\medskip \noindent \textbf{{Quantitative Results.}}
Tab.\ref{tab:remote_result} presents the quantitative registration results on multimodal remote sensing datasets. Our proposed MIFNet, alongside traditional algorithms like OS-SIFT\cite{xiang2018sift} and RIFT2~\cite{li2023rift2}, serves as a general-purpose solution for remote sensing scenarios, requiring no specific fine-tuning. In contrast, CMMNet~\cite{cui2021cross} and ReDFeat~\cite{deng2022redfeat} rely on paired data for training in different scenarios.


We observe that MIFNet consistently enhances cross-modality matching performance across three widely used single-modality image matchers. 
{
In particular, for the Optical-SAR scenario, the matching performance tends to be relatively low for all methods due to substantial nonlinear differences in pixel intensities, severe speckle noise inherent to SAR imaging, and the frequent absence of critical structural details such as road features. Nevertheless, MIFNet significantly improves the Success Rate (SRR) by 24.6\%, 21.9\%, and 17.7\% for SuperPoint, ALIKED, and XFeat, respectively. 
}
Notably, the combination of SuperPoint with MIFNet achieves a registration success rate of 39.7\%, merely 1.7\% lower than the top-performing method, ReDFeat~\cite{deng2022redfeat}. However, ReDFeat is trained on paired multimodal data and lacks generalization to unseen modalities. In contrast, our proposed MIFNet is trained without any exposure to SAR or NIR data.

In the Optical-NIR dataset, which exhibits smaller modality differences, ALIKED and XFeat combined with MIFNet outperform ReDFeat. Specifically, with MIFNet’s enhancement, ALIKED improves by 9.0\% in SRR and 7.8\% in MS, achieving the highest successful registration rate. This demonstrates that our method significantly improves feature matching, thereby enhancing overall registration accuracy. Fig.~\ref{fig:vis_match_remote} shows several visualized feature matching results.

\subsection{Ablation Study}
To justify the effectiveness of the main components as well as the parameter settings in the  proposed MIFNet, we conduct the following ablation studies. 

\begin{figure}[t]
\centerline{\includegraphics[width=0.9\linewidth]{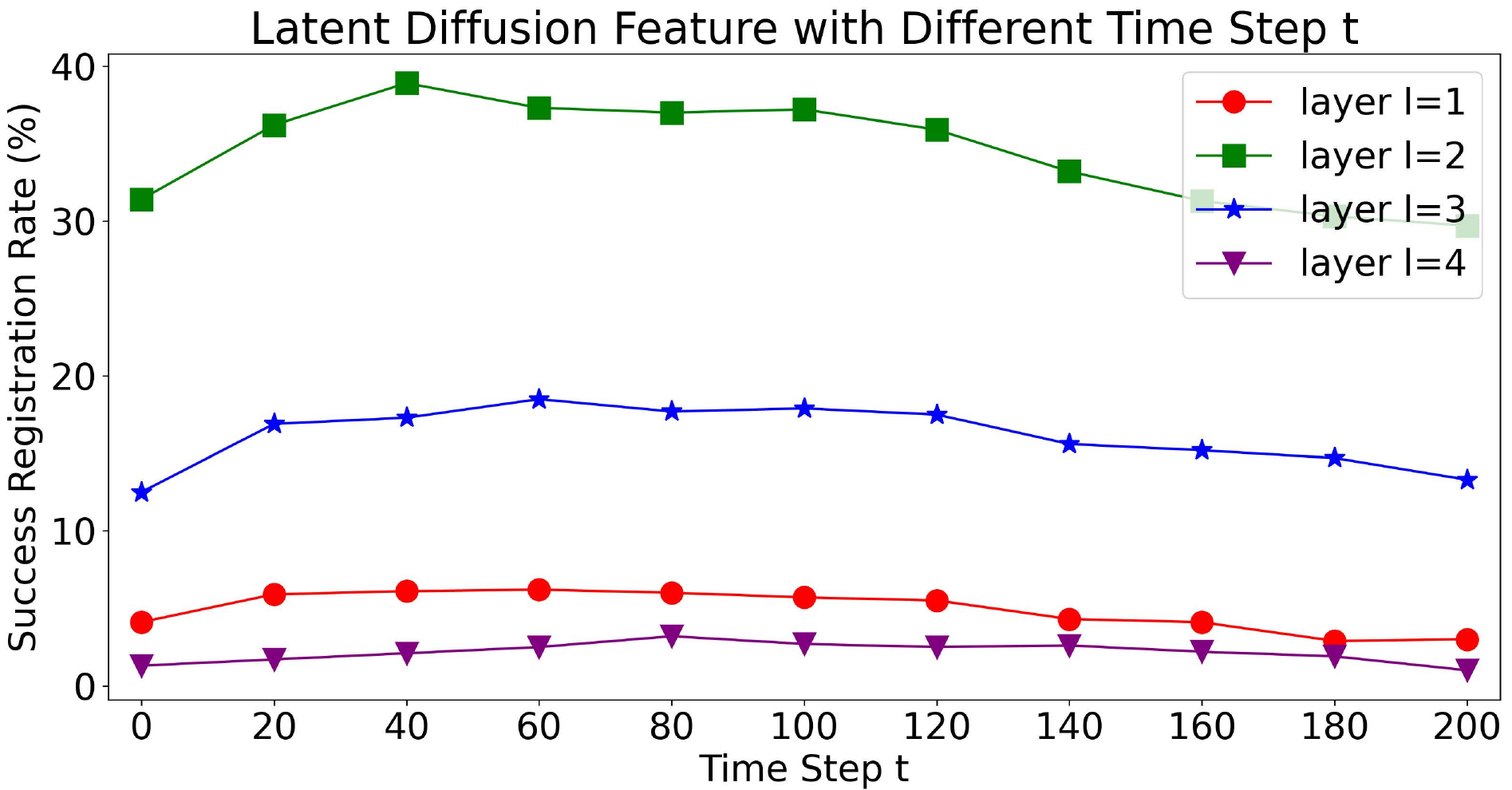}}
\caption{Performance analysis of different hyper-parameters  for latent diffusion features from 
Stable Diffusion v2. $l$ represents the layer of decoder module. $t$ represents  the number of forward noise steps.}
\label{figure:latent_ab}
\end{figure}

\medskip \noindent \textbf{ Hyperparameters of Latent Diffusion Features.}
We analyse two hyper-parameters: 1) the layer $l$ of the
decoder module in Stable Diffusion v2~\cite{rombach2022high}, and 2) the steps  $t$ of forward noise.
As depicted in Fig.~\ref{figure:latent_ab}, the best result is achieved with the setting ($l=2, t=40$). 
{
We observe that features from decoder layer $l=2$ consistently outperform those from shallower ($l=1$) or deeper layers ($l=3$, $l=4$), indicating that intermediate layers better balance semantic abstraction and spatial detail. Additionally, the performance remains stable across a wide range of time steps $t$, demonstrating robustness to latent noise, though a slight drop occurs when $t > 100$ due to weakened semantic coherence. These findings are crucial for designing effective and robust modality-invariant feature learning.
}

\begin{figure}[t]
\centerline{\includegraphics[width=0.90\linewidth]{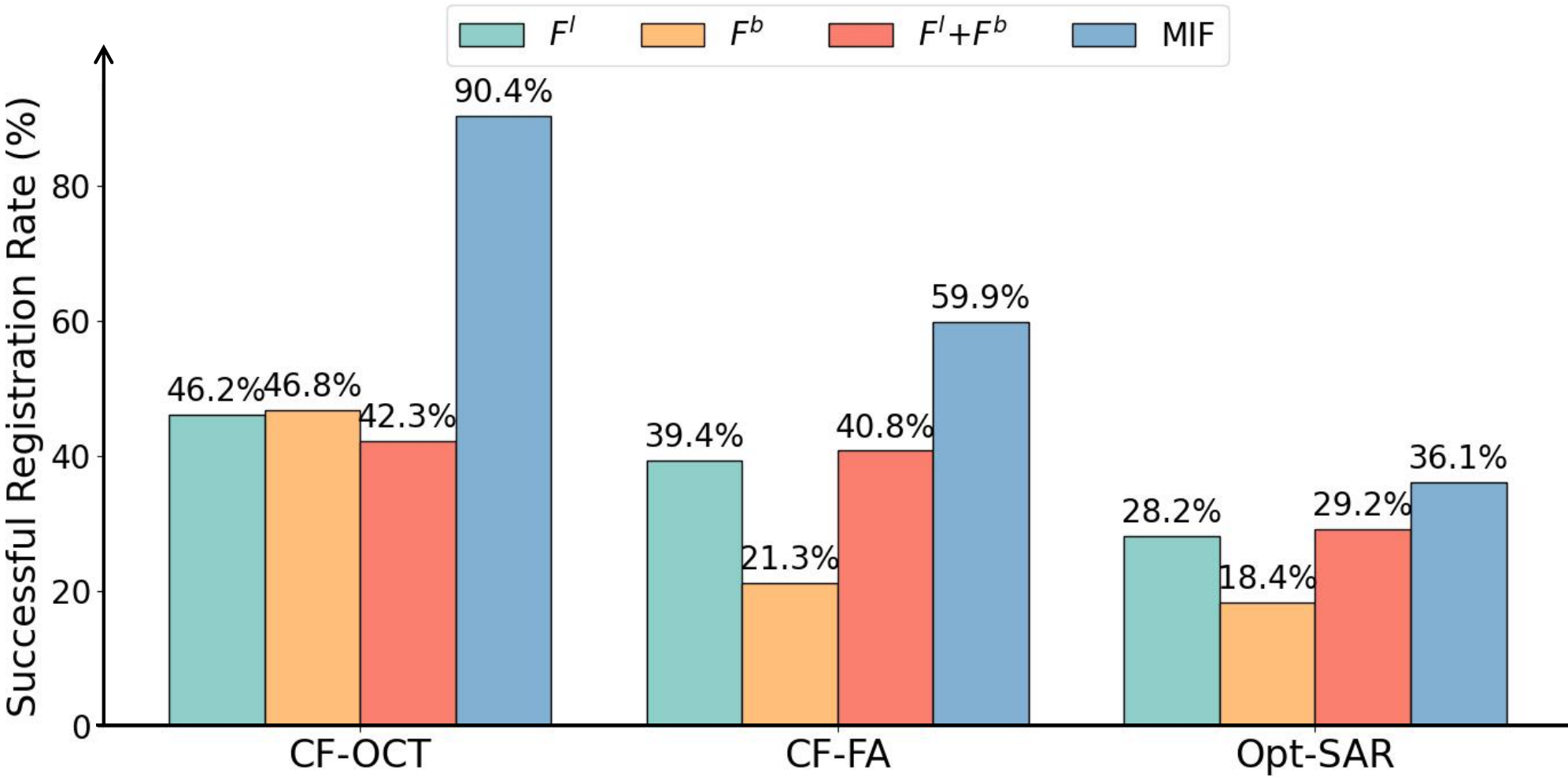}}
\caption{Effectiveness of different feature fusion methods. $F^l$: latent diffusion features. $F^b$: base features of XFeat. $F^l$ + $F^b$: concat the two features.
MIF: the feature of our MIFNet.}
\label{figure:compare_fusion}
\end{figure}

\medskip \noindent \textbf{Performances by Different Feature Fusion Methods.}
We compare the  successful registration  rate of the original latent features and the base features of XFeat~\cite{potje2024xfeat}, as well as their direct concatenation and the fusion using our MIFNet network. The experimental results are shown in Fig.~\ref{figure:compare_fusion}. We found that directly concatenating the two types of features often not only fails to improve performance but can even degrade it in certain scenarios. This is because the latent features and base features exist in different feature spaces, and direct merging leads to poor feature alignment. Our MIFNet outperforms the direct concatenation method in all scenarios, effectively leveraging the strengths of both feature types to the fullest extent.
\renewcommand{\arraystretch}{1.2} 


\begin{table*}[!t]
\centering
\caption{Ablation study on effectiveness of our MIFNet. 
We use XFeat~\cite{potje2024xfeat} as the keypoint detector. $F^{l}$: using the latent diffusion features from stable diffusion v2~\cite{rombach2022high}.  CHA: using Cumulative Hybrid Aggregation module. LFA: using latent feature aggregation module.  }
\label{tab:sem_analysis}
\begin{tabular}{l|ccc|ccc|ccc}
\hline
 \multirow{2}{*}{Configuration}   & \multicolumn{3}{c|}{CF - OCT}    & \multicolumn{3}{c|}{CF - FA}  & \multicolumn{3}{c}{Optical - SAR} \\ 
\cline{2-10}
\multicolumn{1}{l|}{}  & \multicolumn{1}{c}{SRR (\%) $\uparrow$} &\multicolumn{1}{c}{RMSE $\downarrow$}  & \multicolumn{1}{c|}{MAE $\downarrow$}
& \multicolumn{1}{c}{SRR (\%) $\uparrow$} &\multicolumn{1}{c}{RMSE $\downarrow$}  & \multicolumn{1}{c|}{MAE $\downarrow$}
& \multicolumn{1}{c}{SRR (\%) $\uparrow$} &\multicolumn{1}{c}{$H_{\text{err}}$  $\downarrow$}  & \multicolumn{1}{c}{MS (\%) $\uparrow$} \\
\hline
{$F^{l}$} &46.2  &15.7  &42.2  & 39.4 & 21.4  & 53.5 &28.2  & 3.9 & 5.1 \\
{$F^{l}$ + CHA}   & 81.7 & 5.8  & 14.3 & 53.7  & 15.3 & 43.5 & 34.3 & 3.7 & 7.5  \\
{$F^{l}$ + LFA(w/o GMM) + CHA} & 82.9 & 5.2  & 13.6 & 54.5  & 14.1 & 41.3 & 34.5 & 3.7 & 7.6  \\
{{$F^{l}$ + LFA(w/ K-means) + CHA}} & {86.5}  & {4.6} & {12.3} & {55.8} & {13.7} & {37.4} & {35.0} & {3.7} & {7.8}\\
{$F^{l}$ + LFA(w/ GMM) + CHA} & \textbf{90.4}  & \textbf{3.9} & \textbf{10.1} & \textbf{59.9} & \textbf{11.4} & \textbf{35.6} & \textbf{36.1} & \textbf{3.6} & \textbf{8.1}\\
\hline
\end{tabular}
\end{table*}

\medskip \noindent \textbf{Effectiveness of LFA and CHA in MIFNet.} 
To evaluate the effectiveness of the components proposed in MIFNet, we conduct experiments by progressively adding each component to the three cross-modal datasets: CF-OCT, CF-FA, and Optical-SAR. The experimental results are shown in Tab.~\ref{tab:sem_analysis}.
We observe consistent performance improvements across all three test sets. Specifically, aggregating the diffusion and base features using the proposed CHA module enhances the success rate by 35.5\%, 14.3\%, and 6.1\% on the CF-OCT, CF-FA, and Optical-SAR datasets, respectively.
Fusing latent features with base features through multi-layer attention significantly boosts the success rate. While the coarse latent features exhibit semantic invariance, they lack the local geometric details critical for pixel-level matching tasks, such as image matching.

Furthermore, we observed that adding the LFA module without using the GMM clustering to optimize feature distribution did not lead to significant improvements. This is because the latent features output by stable diffusion contain noise, which hinders modality consistency after feature fusion. However, by incorporating the GMM clustering, the semantic distribution of the latent features becomes smoother, resulting in performance improvements of 7.5\%, 5.4\%, and 1.6\% on the three datasets. {We also experimented with K-means clustering as an alternative, but it performed worse than GMM due to its inability to model soft assignments and complex feature distributions.}

\begin{figure}[t]
  \centering
  \subfigure[loss weight $\lambda$]{\includegraphics[width=0.47\linewidth]{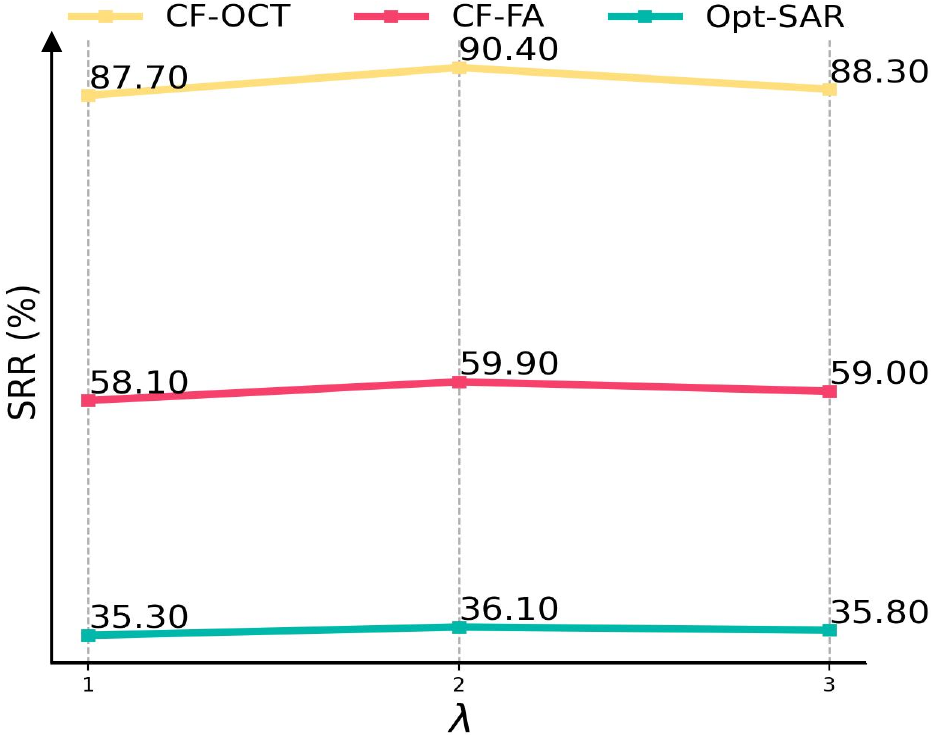} 
  \label{fig:ab_lw}}  
  \subfigure[Number of clusters $K$]{\includegraphics[width=0.47\linewidth]{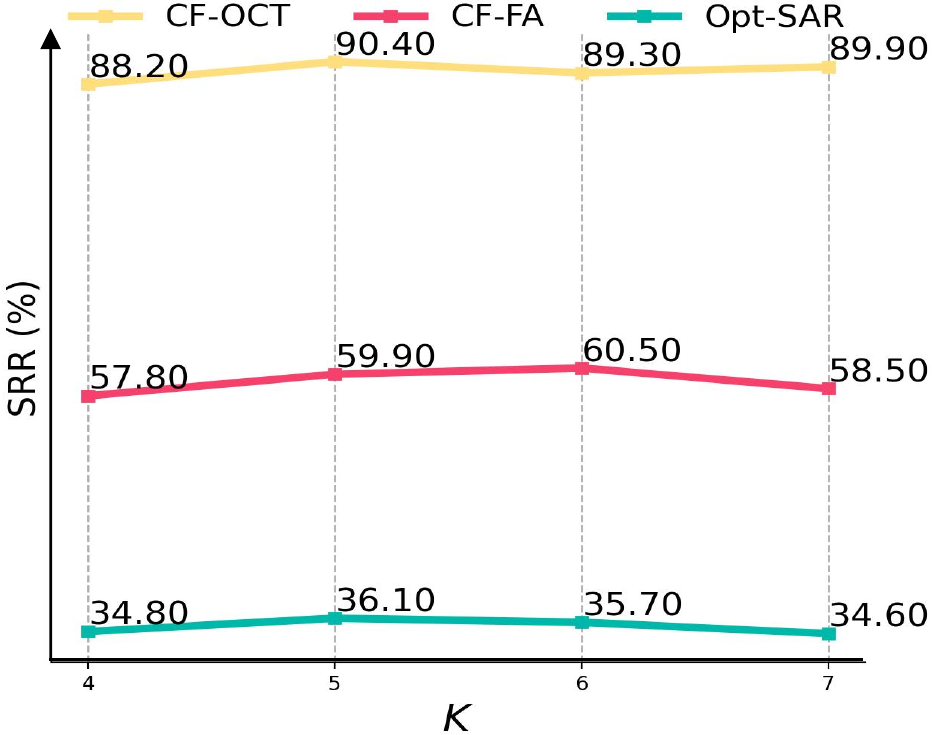} 
  \label{fig:ab_cluster}} 
  \caption{Hyperparameter analysis of MIFNet.}
  \label{fig:abla_hyper}
\end{figure}

\medskip \noindent \textbf{Hyperparameters of MIFNet.}
MIFNet has two hyperparameters, and we conduct experiments using XFeat as the base feature extractor. The first hyperparameter experiment involves the weight parameter \( \lambda \) of the $L_{LFA}$ during training. As shown in Figure~\ref{fig:ab_lw}, we test values of 1.0, 2.0, and 3.0, with the optimal result occurring when \( \lambda=2.0 \). The second hyperparameter is the number of clusters \( K \) in the GMM within the LFA module. If \( K \) is too small, the semantic regions will lack distinctiveness. We test values of \( K = 4, 5, 6, 7 \), and as shown in Figure~\ref{fig:ab_cluster}, the best performance is achieved when \( K = 5 \).


\begin{figure}[t]
  \centering
  \subfigure[Number of attention layers $L$]{\includegraphics[width=0.45\linewidth]{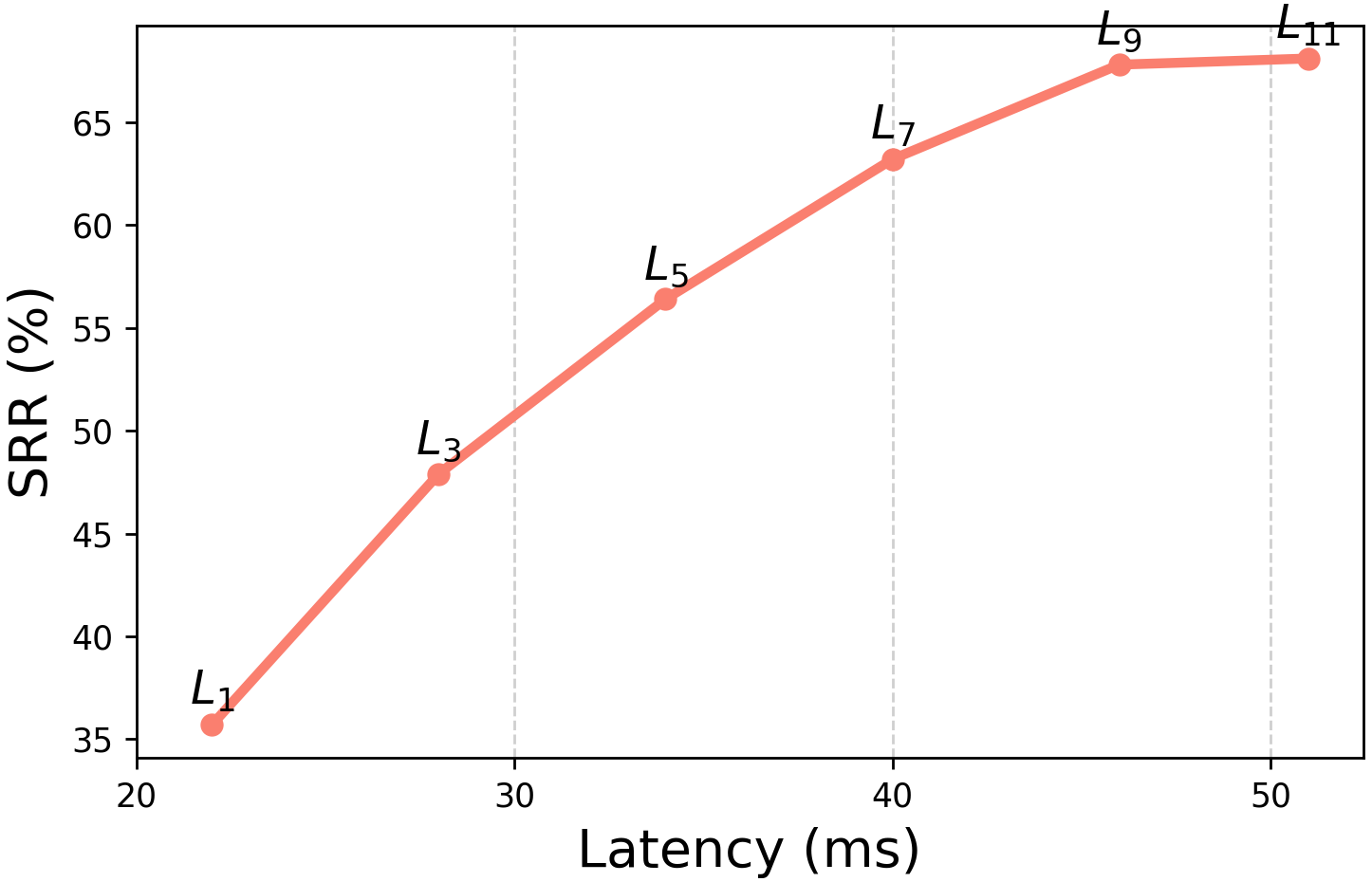}
  \label{figure:ab_param_l}}  
  \subfigure[Foundation model type]{\includegraphics[width=0.48\linewidth]{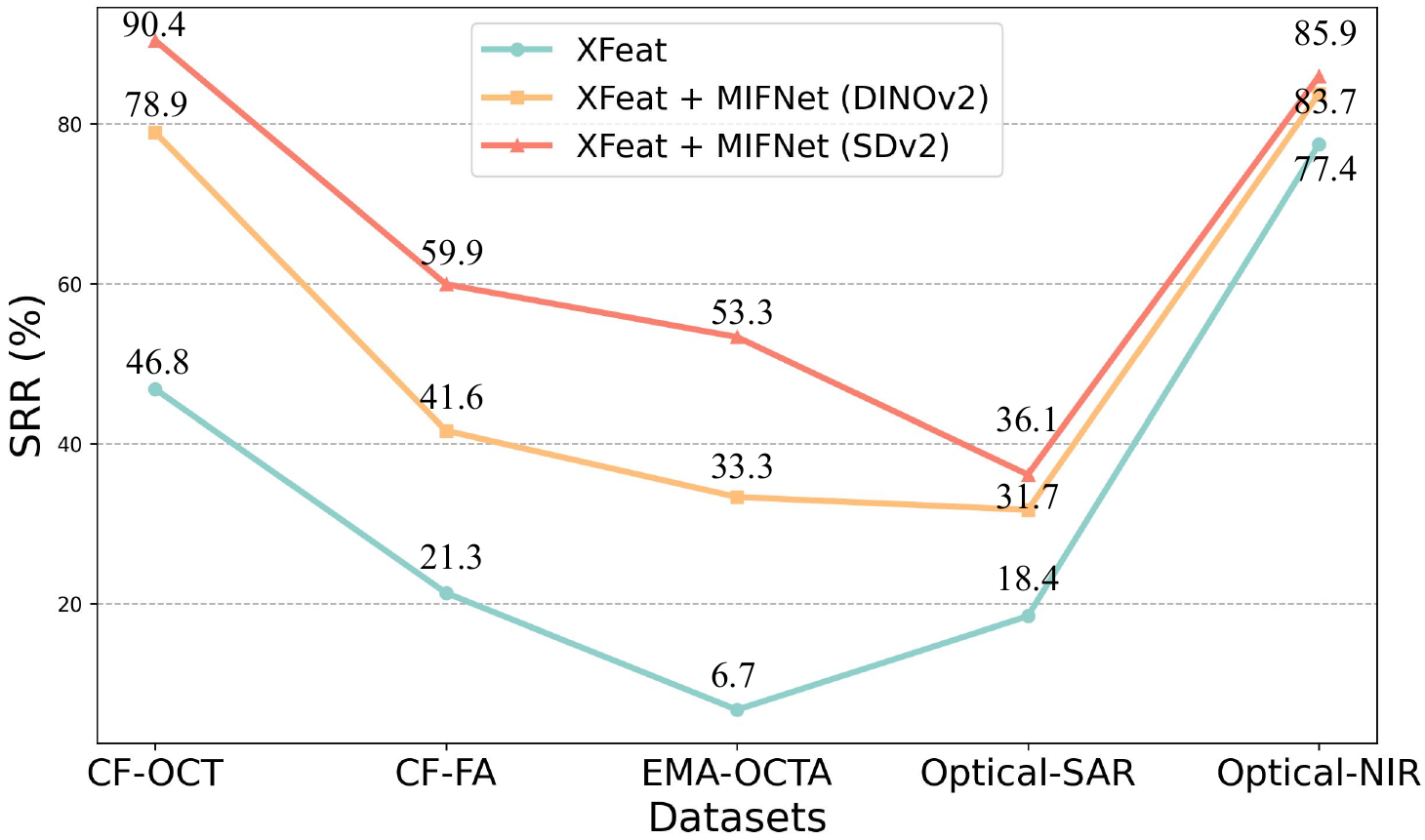}
  \label{figure:ab_foundation}} 
  \caption{{Ablation study of MIFNet on (a) the number of attention layers and (b) different foundation models (DINOv2 vs. Stable Diffusion v2) using XFeat as the base matcher.}}
  \label{fig:abla_attention_foundation}
\end{figure}

\medskip \noindent \textbf{ Influence of the Number of Attention Layers.}
{
We evaluate the impact of the number of attention layers in MIFNet for \( L \in \{1, 3, 5, 7, 9, 11\} \) and report the average registration success rate (SRR) on three fundus datasets: CF-OCT, CF-FA, and EMA-OCTA. XFeat~\cite{potje2024xfeat} is used as the base feature extractor.
As shown in Fig.~\ref{figure:ab_param_l}, increasing $L$ from 1 to 9 improves performance. The results with \( L = 9 \) and \( L = 11 \) are comparable. We choose \( L = 9 \) as the default setting to maintain a good balance between accuracy and inference time.}

\medskip \noindent \textbf{ Influence of Different Foundation Models.}
{
We compare two widely-used pretrained foundation models, Stable Diffusion v2 (SDv2) and DINOv2, to investigate their impact on MIFNet's cross-modal matching performance.
}

{
As shown in Fig.~\ref{figure:ab_foundation}, both models improve upon the original XFeat baseline across all datasets, with SDv2 consistently outperforming DINOv2. In the fundus imaging scenario, SDv2 achieves higher SRR than DINOv2 by 11.5\%, 18.3\%, and 20.0\% on the CF-OCT, CF-FA, and EMA-OCTA datasets, respectively. The performance gap is more pronounced in datasets with greater modality discrepancies. In the remote sensing scenario, the improvement margins are relatively smaller, with gains of 4.4\% on Optical-SAR and 2.2\% on Optical-NIR, likely due to the higher visual similarity between remote sensing images and the natural images used during pretraining. Compared to DINOv2, SDv2 produces more modality-invariant features in zero-shot settings, owing to its stronger semantic representation capability. These results suggest that integrating even more powerful foundation models could further improve the overall performance of MIFNet.
}

\begin{table}[!t]
\centering
\caption{{Influence of training on different modalities for multimodal image matching. MIFNet(O): trained on optical images, MIFNet(S): trained on SAR images, MIFNet(O+S): trained on a mixed dataset composed of half optical and half SAR images.}}
\label{tab:ab_dataset_influence}
\begin{tabular}{l|cc|cc}
\hline
\multicolumn{1}{l|}{\multirow{2}{*}{Methods}} & \multicolumn{2}{c|}{Optical - SAR}   & \multicolumn{2}{c}{Optical - NIR}    \\ 
\multicolumn{1}{l|}{}   
& \multicolumn{1}{c}{${\text{SRR} (\%)}$ $\uparrow$} 
& \multicolumn{1}{c|}{MS (\%) $\uparrow$ } &
\multicolumn{1}{c}{${\text{SRR} (\%)}$ $\uparrow$} 
& \multicolumn{1}{c}{MS (\%) $\uparrow$ }\\
\hline 
\multicolumn{1}{l|}{XFeat}  & 18.4   & 3.0 & 77.4  & 37.8  \\
\multicolumn{1}{l|}{MIFNet (O)} &  36.1  & 8.1 & \textbf{85.9}  & \textbf{46.1} \\
\multicolumn{1}{l|}{MIFNet (S)} &  \textbf{39.3}  & \textbf{9.2} & 83.0  & 45.3 \\
\multicolumn{1}{l|}{MIFNet (O+S)} &  37.5  & 8.7 & {85.5}  & {45.9} \\
\hline
\end{tabular}
\end{table}
\medskip \noindent \textbf{Influence of Training Dataset Composition.}
{
Our approach is trained using synthetic image pairs generated by applying random geometric transformations to individual images, without requiring any paired multimodal data.  To assess the effect of training data composition, we conduct experiments with three configurations: optical only (O), SAR only (S), and a mixed set of unpaired optical and SAR images (O+S).  Each single-modality dataset contains 2,400 images, while the mixed set includes 1,200 images from each modality.
}

{
As shown in Tab.~\ref{tab:ab_dataset_influence}, all three training configurations outperform the baseline (XFeat). In  Optical-SAR scenario, SAR-only training achieves the best SRR (39.3\%). The mixed dataset also improves performance by 1.4\% over Optical-only. In contrast, in  Optical-NIR scenario, SAR-only training yields the lowest performance (83.0\%). The mixed dataset slightly underperforms compared to optical-only. These results suggest that incorporating data from modalities similar to the test domain can improve performance. Nevertheless, since optical data are widely available and already effective, we adopt optical-only training in all other experiments.
}


\begin{figure}[t]
\centerline{\includegraphics[width=0.95\linewidth]{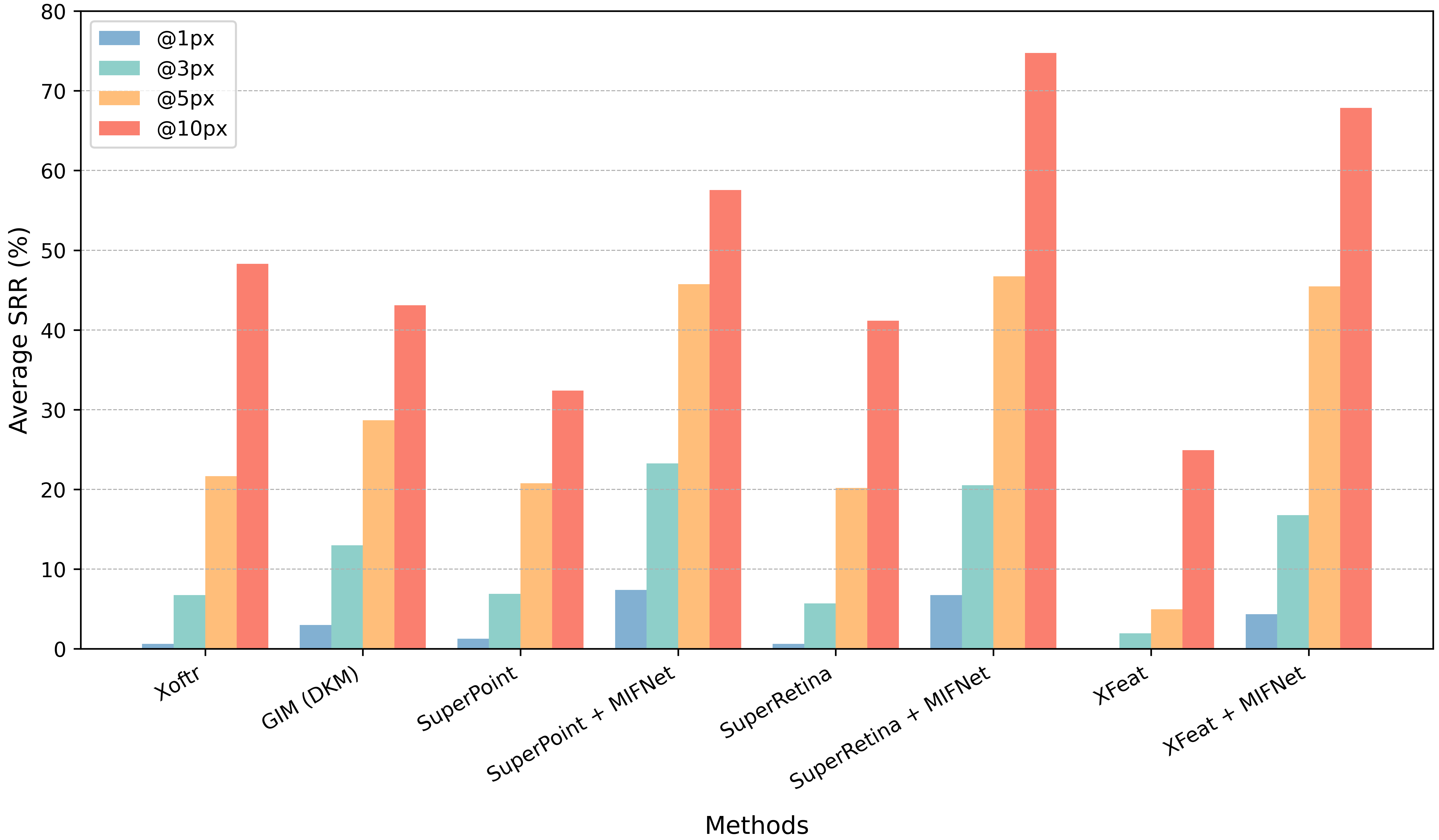}}
\caption{{Average successful registration  rate (SRR) across three fundus datasets (CF-OCT, CF-FA, EMA-OCTA) under different RMSE thresholds \{1, 3, 5, 10\} pixels.}
}
\label{figure:rmse_variance}
\end{figure}

\medskip
\noindent
\textbf{Effect of Varying RMSE Thresholds.}  
{
To further assess the robustness of our MIFNet, we report the successful registration rate (SRR) under three commonly used RMSE thresholds: 1, 3, 5, and 10 pixels. As shown in Fig.~\ref{figure:rmse_variance}, our proposed MIFNet improves the robustness of baseline detectors such as SuperPoint~\cite{detone2018superpoint}, SuperRetina~\cite{liu2022semi}, and XFeat~\cite{potje2024xfeat} across all thresholds. Moreover, when combined with SuperRetina, MIFNet consistently outperforms other baselines, including Xoftr~\cite{tuzcuouglu2024xoftr} and GIM~\cite{xuelun2024gim}, under different thresholds.
}

\subsection{Runtime Analysis and Computational Complexity}

\begin{figure}[t]
\centerline{\includegraphics[width=0.85\linewidth]{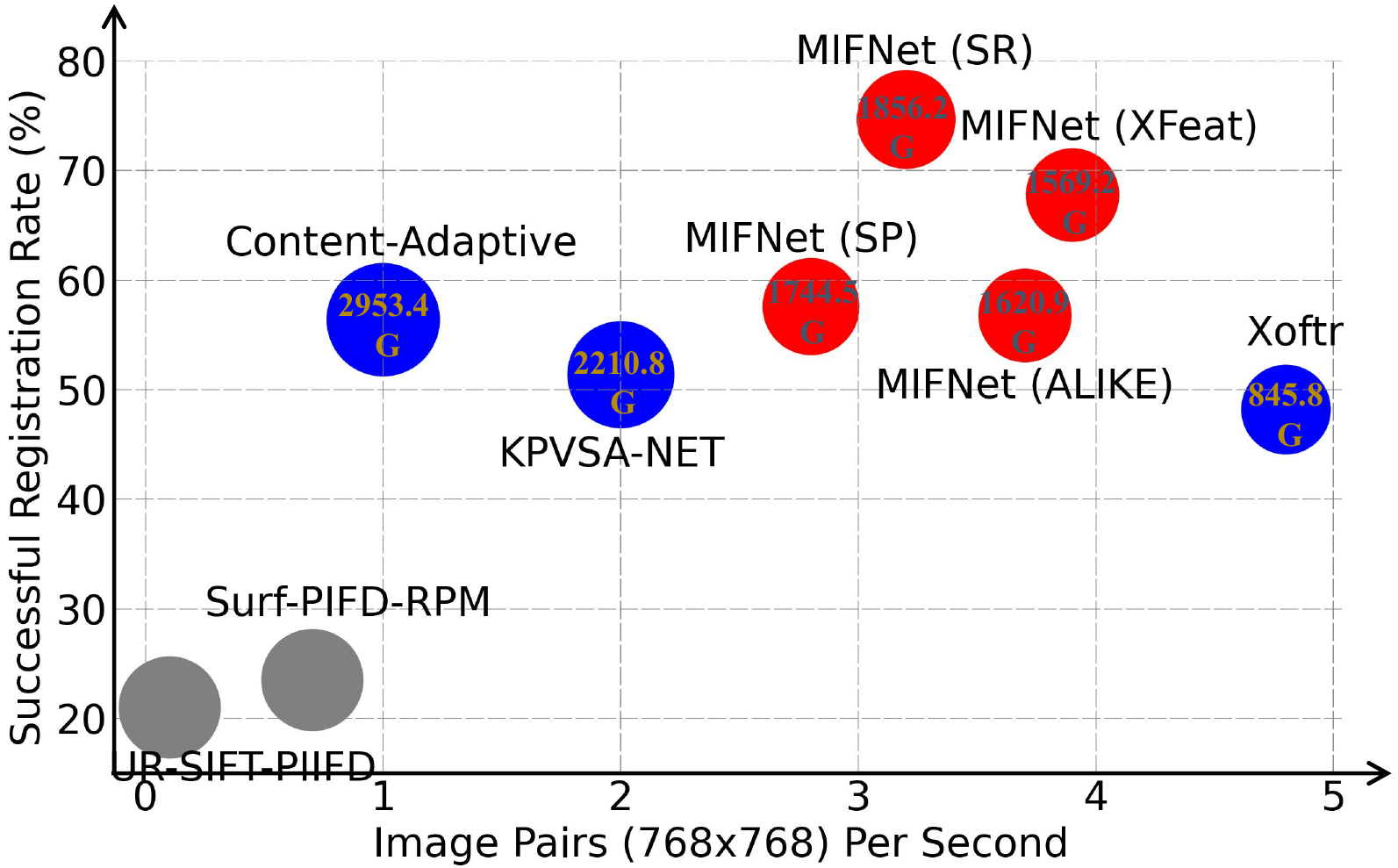}}
\caption{Runtime analysis. We computed the average SRR and runtime of various methods on CF-OCT, CF-FA, and EMA-OCTA. Red: Methods that do not require cross-modal data for training. Blue: Methods that utilize data from different modalities for training. {The number inside each circle represents the computational cost in FLOPs.}
}
\label{figure:runtime_analyse}
\end{figure}


We conduct experiments on a system equipped with an NVIDIA 3090 GPU (24 GB) and an 8-core Intel(R) Xeon(R) Platinum 8375C CPU. {The runtime and computational complexity are important metrics for applying image matching to downstream tasks.} The average  successful rate and runtime are reported across three retinal datasets. {Both runtime and GFLOPs are measured as the total latency and computational cost for registering one image pair.} Fig.~\ref{figure:runtime_analyse} presents the results of various methods. Among these, UR-SIFT-PIFD~\cite{zhang2018automated} and SURF-PIFD-RPM~\cite{wang2015robust} are traditional algorithms that do not use GPU resources.  {Content-Adaptive~\cite{wang2021robust}, KPVSA-NET~\cite{sindel2022multi}, and Xoftr~\cite{tuzcuouglu2024xoftr} are learning-based multimodal methods that rely on paired cross-modal training data}. 
In contrast, MIFNet requires only training on color fundus images and performs zero-shot testing on unseen multimodal image pairs. 
Specifically, MIFNet combined with SuperRetina achieves the highest success rate and significantly outperforms traditional algorithms in terms of speed, while being only marginally slower than Xoftr.
 {The main computational cost comes from Stable Diffusion v2, which is used during inference for latent feature extraction.  SuperRetina + SDv2 + MIFNet  requires a total of 1856.2 GFLOPs, with SuperRetina accounting for 301.2 GFLOPs and SDv2 for 1522.7 GFLOPs. In comparison, MIFNet's network layers only add 32.3 GFLOPs and 11.3M parameters, making it lightweight and efficient.}

\begin{figure}[!t]  
    \centering
\subfigure{\includegraphics[width=0.8in]{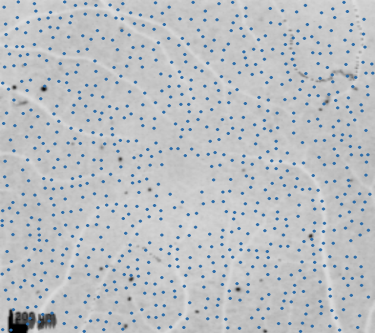}}
\subfigure{\includegraphics[width=0.8in]{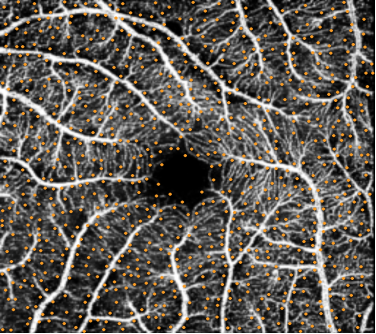}}    
\subfigure{\includegraphics[width=0.8in]{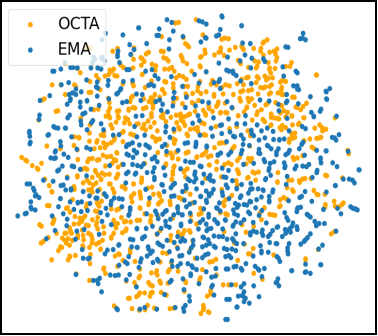}}    
\subfigure{\includegraphics[width=0.8in]{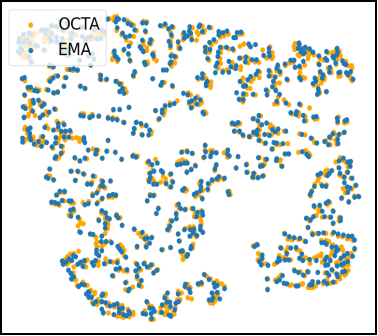}}\setcounter{subfigure}{0}
\subfigure[]{\includegraphics[width=0.8in]
{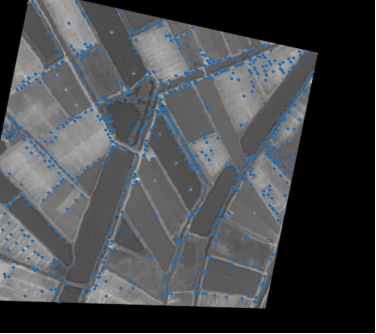}}    
\subfigure[]{\includegraphics[width=0.8in]{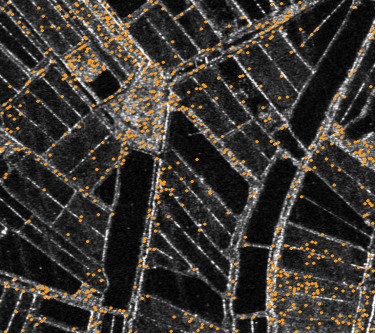}}    
\subfigure[]{\includegraphics[width=0.8in]{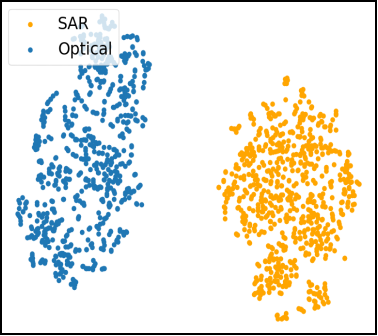}}    
\subfigure[]{\includegraphics[width=0.8in]{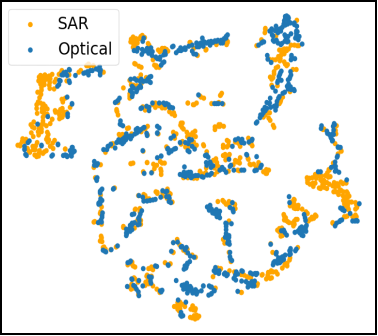}}    
    \caption{t-SNE visualization of features   from multimodal image pairs. As shown, MIFNet is able to reduce the cross modality variance for final keypoint descriptor from the multimodal data. (a)-(b) Paired  EMA-OCTA (first row) and   Optical-SAR (second row) images with detected keypoints. (c) Features of keypoints by XFeat. (d) Features of keypoints by MIFNet. }
\label{fig:discuss_invariant}
\end{figure}

\section{Discussions}
To better understand how MIFNet improves the {geometric invariance}, modality invariance, distinctiveness, {and generality of the learned features}, we conduct four analyses in this section. 
Without losing generality, we employ XFeat~\cite{potje2024xfeat} as the feature extractor.

\begin{table}[!t]
\centering
\caption{{Successful registration rate under rotation perturbations.}}
\label{tab:ab_rotation_influence}
\setlength{\tabcolsep}{4pt} 
\begin{tabular}{l|ccccc}
\hline
\multicolumn{1}{l|}{\multirow{1}{*}{Methods}} & \multicolumn{1}{c}{CF-OCT} & \multicolumn{1}{c}{CF-FA} & \multicolumn{1}{c}{EMA-OCTA} & \multicolumn{1}{c}{Opt-SAR} & \multicolumn{1}{c}{Opt-NIR}  \\ 

\hline 
\multicolumn{1}{l|}{Xoftr}      & 48.1 & 41.3  & 6.7     & 19.8  &  45.0 \\
\multicolumn{1}{l|}{GIM}        & 73.1 & 36.5  & 0  & 6.8  & 48.5  \\
\multicolumn{1}{l|}{SuperPoint} & 57.7 & 8.8   & 0     & 2.5  & 28.5      \\
\multicolumn{1}{l|}{SP+MIFNet}  & \cellcolor{gray!10} 86.5 & \cellcolor{gray!10} 60.0  & \cellcolor{gray!10} 13.3  & \cellcolor{gray!10} 20.0 & \cellcolor{gray!10} 53.5 \\
\multicolumn{1}{l|}{ALIKED}     & 15.4 & 6.6   & 0     & 1.3  &  30.5 \\
\multicolumn{1}{l|}{AL+MIFNet}  & \cellcolor{gray!10} 69.2 & \cellcolor{gray!10} 48.8  & \cellcolor{gray!10} 26.7  & \cellcolor{gray!10} 18.5  & \cellcolor{gray!10} 55.0 \\
\multicolumn{1}{l|}{XFeat}      & 17.3 & 3.5   & 0     & 4.0  & 31.0 \\
\multicolumn{1}{l|}{XF+MIFNet}  & \cellcolor{gray!10} 88.5 & \cellcolor{gray!10} 57.5  & \cellcolor{gray!10} 46.7  & \cellcolor{gray!10} 28.3  & \cellcolor{gray!10} 68.5 \\
\hline
\end{tabular}
\end{table}

\subsection{Geometric Invariance Analysis}  
{
To evaluate the robustness of our method under rotational variations, we apply random rotations to the input image pairs: fundus images are rotated within \([-36^\circ, 36^\circ]\), and remote sensing images within \([-90^\circ, 90^\circ]\). 
As shown in Tab.~\ref{tab:ab_rotation_influence}, the performance of lightweight single-modality detectors such as SuperPoint, ALIKED, and XFeat degrades significantly under increased rotation. In contrast, our method exhibits strong robustness to such geometric transformations. Notably, when combined with XFeat, our method achieves the best performance across all test datasets under rotation conditions.
}

\subsection{Modality Invariance Analysis}
We begin by investigating feature extraction on the EMA-OCTA and Optical-SAR datasets, which represent challenging multimodal image matching scenarios with significant nonlinear intensity differences in retinal and remote sensing images, respectively. To ensure a comprehensive evaluation, we sample distributed keypoints across two randomly selected image pairs and apply the t-SNE algorithm for dimensionality reduction to visualize the features in two-dimensional space. Fig.~\ref{fig:discuss_invariant} displays paired image data with detected keypoints and the corresponding features. As shown in the figure, we observe better overlap of points in (d) compared to (c), indicating improved cross-modal feature invariance. In the particularly challenging Optical-SAR scenario, where the original feature distances are substantial, our algorithm maintains strong modality invariance for most points. These results highlight the robustness of our approach in aligning features across modalities, even under extreme intensity variations.

\subsection{Distinctiveness Analysis}
\begin{figure}[!t]
  \centering
  \subfigure[]{\includegraphics[width=0.95\linewidth]{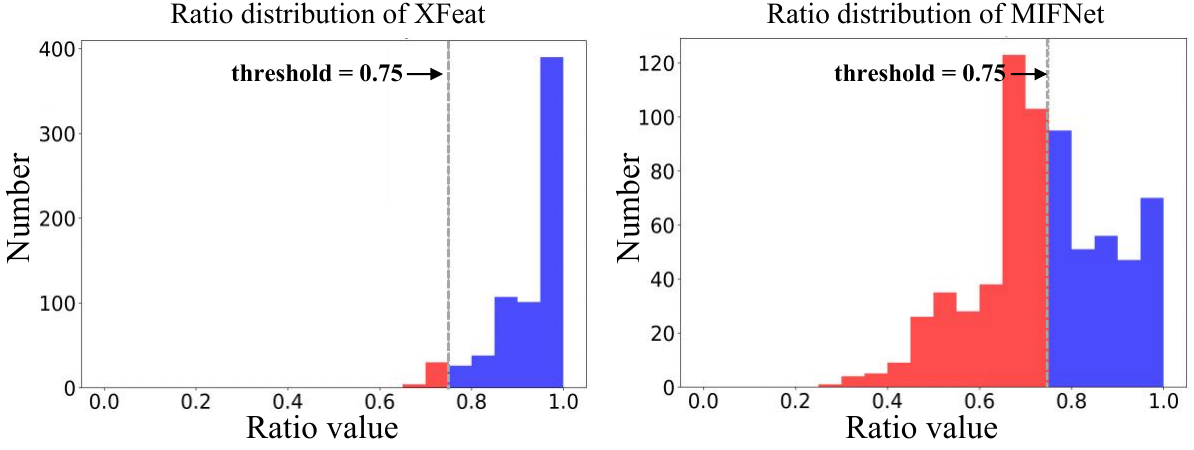}
  \label{fig:mot-a}}  
  \subfigure[]{\includegraphics[width=0.95\linewidth]{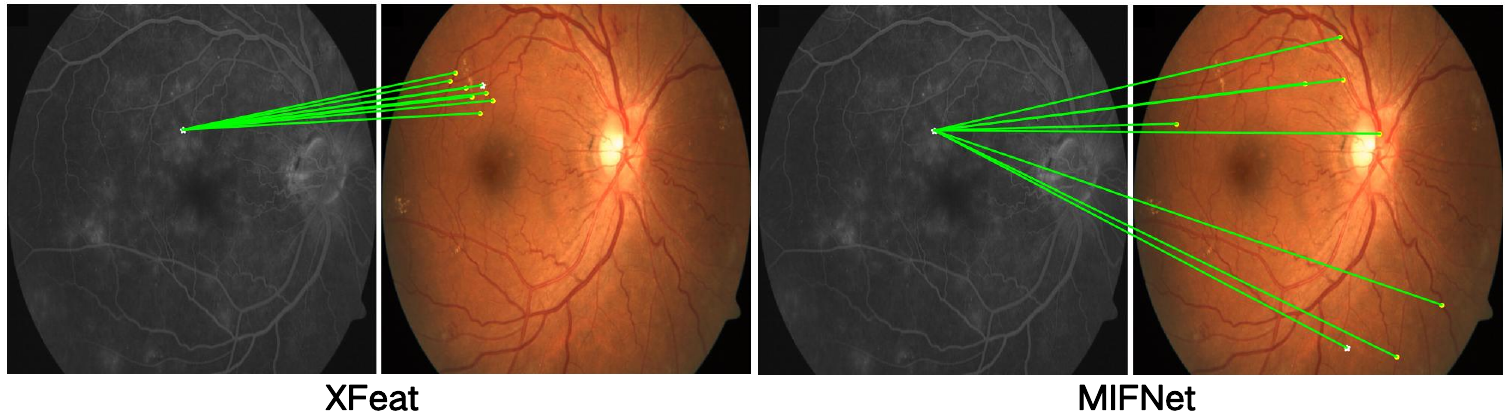}
  \label{fig:mot-b}} 
  \caption{Discriminative Analysis of Features for XFeat~\cite{potje2024xfeat} and MIFNet on CF-FA Image Pair. (a) Distribution of Lowe's Ratio~\cite{david2004distinctive}. (b) A randomly selected query point in the FA image and its eight most similar keypoints in the CF image.
  }
  \label{fig:images_motivation}
\end{figure}

We also investigate the distinctiveness of the modality-invariant features. Distinctiveness refers to how well the description of a keypoint is differentiated from other unpaired keypoints, reducing erroneous matches. To evaluate this, we adopt the commonly used Lowe's ratio test~\cite{david2004distinctive}, where a smaller ratio indicates better descriptor recognizability.
Fig.~\ref{fig:mot-a} shows the ratio distribution of all matched pairs. The histogram demonstrates that MIFNet outperforms XFeat, with more descriptor pairs having ratios below 0.75, indicating that our method achieves stronger distinctiveness in cross-modality scenarios.

We further compare how feature quality affects matching. In Fig.\ref{fig:mot-b}, we randomly select a query keypoint and its eight most similar keypoints. For XFeat\cite{potje2024xfeat}, the eight closest keypoints are scattered across various regions. In contrast, the eight most similar keypoints identified by MIFNet are concentrated in a small region. This observation explains why MIFNet outperforms XFeat in matching performance.

\begin{table}[t]
\centering \tiny
\caption{{
Generalization results on the FIRE dataset (color fundus–color fundus) and the SRIF dataset (optical–optical). The MIFNet models used here are trained on the FA modality and the SAR modality, respectively.
}} 
\label{tab:discuss_generable}
\resizebox{1.0\linewidth}{!}{
\begin{tabular}{l|cc|cc}
\hline
\multirow{2}{*}{Methods} & \multicolumn{2}{c|}{FIRE (CF–CF)} & \multicolumn{2}{c}{SRIF (Opt–Opt)} \\
\cline{2-5}
 & SRR(\%) $\uparrow$ & RMSE $\downarrow$ & SRR(\%) $\uparrow$ & $H_{\text{err}}$ $\downarrow$ \\
\hline
ALIKED             & 85.0 & 13.4 & 32.5 & 3.3 \\
ALIKED + MIFNet    & 95.1 & 5.1  & 48.5 & 2.6 \\
XFeat              & 91.7 & 8.0  & 38.0 & 3.1 \\
XFeat + MIFNet     & \textbf{97.7} & \textbf{3.9} & \textbf{52.5} & \textbf{2.5} \\
\hline
\end{tabular}
}
\end{table}

\subsection{Generalization Capability Across Different Modalities}
{
To evaluate the cross-modal generalization capability of our MIFNet, we conduct additional experiments where the model is trained on one modality and tested on image pairs from another different modality.
In the fundus imaging scenario, MIFNet is trained using 400 fluorescein angiography (FA) images and evaluated on the FIRE dataset~\cite{hernandez2017fire}, which contains 134 color fundus (Color Fundus–Color Fundus) image pairs. In the remote sensing scenario, the model is trained on 2,400 SAR images and tested on Optical–Optical image pairs from the SRIF dataset~\cite{li2023multimodal}.
}

{
As shown in Tab.~\ref{tab:discuss_generable}, even though both the training and test data for these baselines come from similar modalities, our method still improves the performance of baseline methods (ALIKED~\cite{zhao2023aliked}, XFeat~\cite{potje2024xfeat}).
This highlights the strong generalization ability of MIFNet, which effectively improves the matching performance of baseline models across different modalities.
}

\begin{figure}[t]
\centerline{\includegraphics[width=1.0\linewidth]{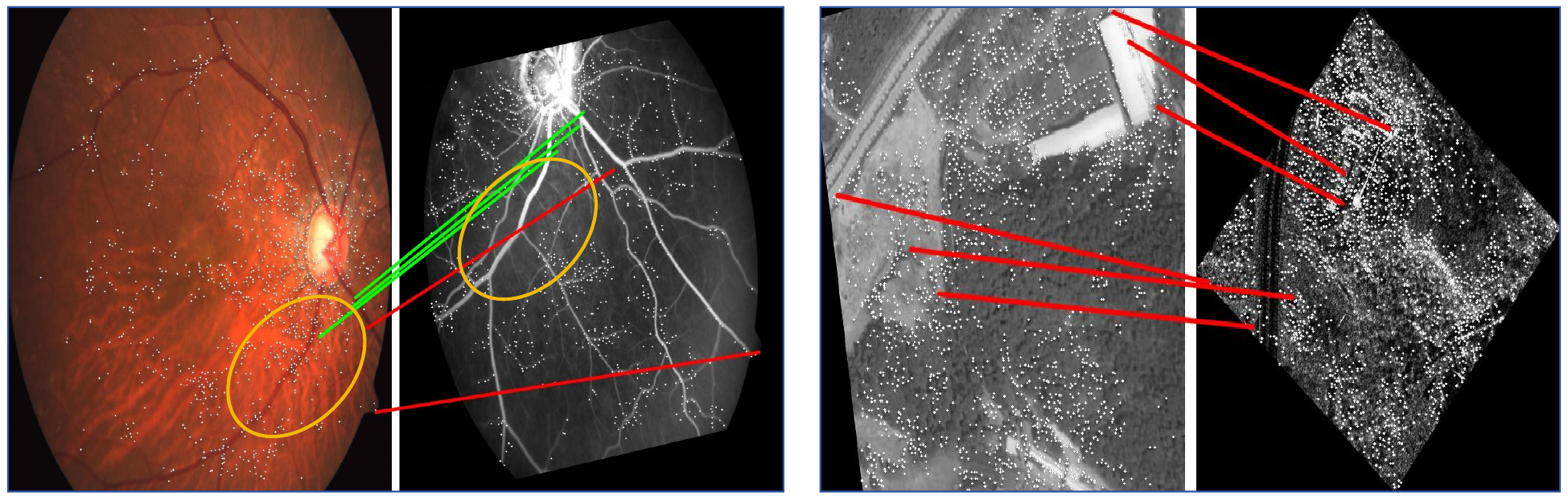}}
\caption{
{
Visualization of some failure cases. 
}}
\label{fig:limitation}
\end{figure}

\subsection{Failure Case Analysis}
{
Although our method achieves strong performance across various scenarios, it still faces challenges in certain cases. As shown in Fig.~\ref{fig:limitation}, on the left, large cross-modal appearance gaps lead to poor keypoint repeatability, which limits our method since it mainly enhances descriptor robustness. On the right, heavy noise and lack of semantic structure (e.g., in degraded SAR images) also affect matching accuracy. Addressing these issues will be part of our future work.
}

\section{Conclusion} 
In this paper, we propose an effective modality-invariant feature learning network (MIFNet) for multimodal image matching, leveraging latent diffusion features derived from the diffusion model. For that, we design a novel self-supervised framework to effectively align and aggregate latent diffusion feature with the base feature from existing single-modality keypoint detection and description. This gives rise to modality-invariant features, which generalize well to unseen modalities. Extensive experiments on three multimodal retinal  datasets and two remote-sensing datasets demonstrate the superiority and scalability of our approach. In the future, we would like to further improve the latency of our MIFNet.

\section{Acknowledgment}
This work was supported in part by the National Key Research and Development Program of China (Grant No. 2023YFC2705700), the National Natural Science Foundation of China (Grant Nos. 62222112, 62225113, and 62176186), the Innovative Research Group Project of Hubei Province (Grant No. 2024AFA017), and the Key Research and Development Program of Zhejiang Province (Grant No. 2024C03204). This project was also supported by A*STAR under the Coastal Protection and Flood Management Research Programme (Grant No. 2024/T2/H1-P4).


 

\bibliographystyle{IEEEtran}
\bibliography{tip}

\end{document}